# Towards Vision Zero: The TUM Traffic Accid3nD Dataset

Walter Zimmer [1✉], Ross Greer [2], Xingcheng Zhou [1], Rui Song [1,4], Hu Cao [1],
Daniel Lehmberg [1], Marc Pavel [1], Ahmed Ghita [3], Akshay Gopalkrishnan [5],
Holger Caesar [6], Mohan Trivedi [5], Alois C. Knoll [1]

[1] Technical University of Munich, [2] University of California Merced, [3] SETLabs Research GmbH,
[4] Fraunhofer IVI, [5] University of California San Diego, [6] Delft University of Technology

accident-dataset.github.io

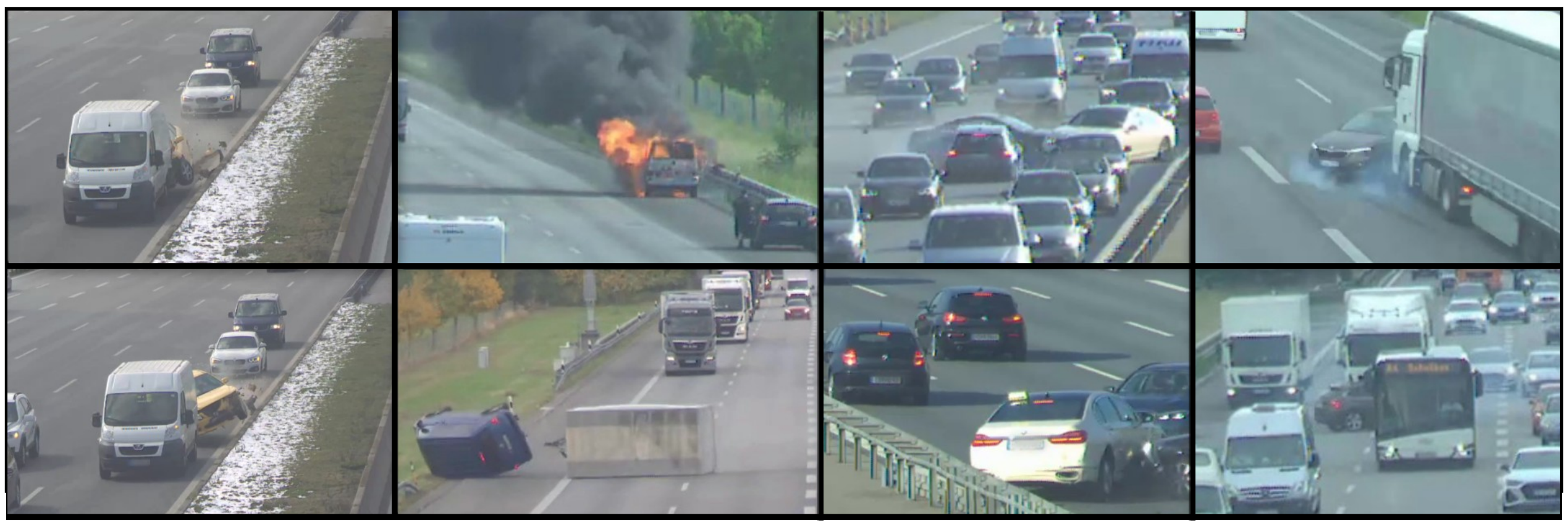

Figure 1. **Raw TUMTraf-Accid3nD dataset.** Accidents are recorded from roadside cameras on the A9 Test Field for Autonomous Driving in Munich, Germany. The dataset includes scenes with high-speed traffic, collisions, and overturning vehicles. Some vehicles are catching fire.

## Abstract

*Even though a significant amount of work has been done to increase the safety of transportation networks, accidents still occur regularly. They must be understood as unavoidable and sporadic outcomes of traffic networks. No public dataset contains 3D annotations of real-world accidents recorded from roadside camera and LiDAR sensors. We present the TUM Traffic Accid3nD (TUMTraf-Accid3nD) dataset, a collection of real-world highway accidents in different weather and lighting conditions. It contains vehicle crashes at high-speed driving with 2,634,233 labeled 2D bounding boxes, instance masks, and 3D bounding boxes with track IDs. In total, the dataset contains 111,945 labeled image and point cloud frames recorded from four roadside cameras and LiDARs at 25 Hz. The dataset contains six object classes and is provided in the OpenLABEL format. We propose an accident detection model that combines a rule-based approach with a learning-based one. Experiments and ablation studies on our dataset show the robustness of our proposed method. The dataset, model, and code are available on our website.*

## 1. Introduction

Vision Zero is a worldwide initiative to reduce road deaths and serious injuries through innovative, data-driven interventions. Autonomous driving (AD) and intelligent infrastructure play a significant role in making roads safer by preventing accidents before they happen. Collecting data on events that occur most rarely (*long-tail events*) is important for robust machine learning of data-driven perception, planning, and control models in robotic systems [1]. In the case of AD, however, these events are especially costly to collect [2, 3, 4]. Long-tail events such as accidents and near-misses [5] come at great risk to human life and are otherwise difficult to stage and capture in the natural driving environment [6, 7]. The time between an accident and the arrival of medical assistance significantly impacts whether the passengers of a vehicle survive an accident. Automatic accident detection reduces this time and has the potential to save lives.

Toward the goal of Vision Zero, we propose *TUMTraf-Accid3nD*, a 3D perception dataset specifically curated for accident scenarios. We present novel labeling methods alongside newly introduced tasks, including 3D object detection, tracking, and accident detection. Furthermore, we demonstrate how vision-language models can enhance safety analysis to improve overall situational awareness. Our dataset offers 3D accident labels with tracking information. It supports the development of safer autonomous systems through multiple directions: (1) the data supports improved learning on a variety of perception-related tasks, such as detection [8, 9, 10, 11, 12, 13, 14, 15, 16, 17, 18, 19], tracking, and segmentation [20]. (2) The nature of the dataset allows for the study of methods of cooperative perception between roadside sensors observing the same scene from different view angles. Cooperative perception allows for a reduction of occlusion through shared information. (3) Roadside sensors allow the creation of digital twins of the traffic scene [21, 22, 23, 24]. These digital twins expand the visibility range beyond the egocentric view, which may be used to provide adequate warning lead time for safe approaches and control transitions for risky traffic events [25].

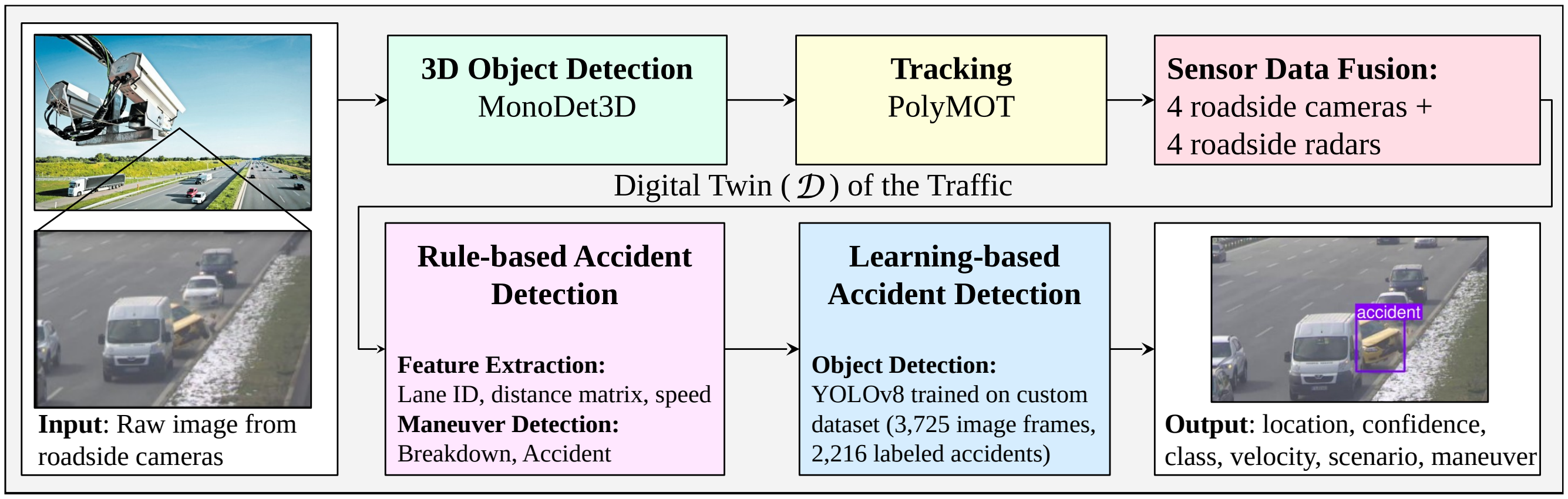

Figure 2. **Accident detection pipeline.** We use advanced 3D perception techniques and multi-sensor data fusion to create a real-time digital twin of the traffic. Starting with raw camera images, the framework first performs 3D object detection using MonoDet3D to identify and localize vehicles in three dimensions. Following detection, Poly-MOT tracking is applied to maintain continuity across frames, while sensor data fusion combines inputs from four roadside cameras and four radars. The digital twin is then used in two accident detection modules: 1) The **Rule-based Accident Detection** module extracts features such as lane IDs, distance matrices, and velocities, identifying potential accidents through predefined maneuver detection rules. 2) The **Learning-based Accident Detection** module employs a YOLOv8 object detector, trained on a custom dataset, to detect accident events. The final output includes the object's location, confidence score, class, velocity, and detected scenario or maneuver.

In addition to standard computer vision tasks—such as 2D object detection, instance segmentation, 3D object detection, sensor data fusion, tracking, trajectory prediction, and accident classification, our dataset also supports accident analysis. This involves reconstructing the sequence of events to understand how the accident occurred and what contributing factors may have played a role. For instance, was the vehicle speeding? Was there a traffic jam a short distance ahead? Did the driver appear inattentive or react too late?

Furthermore, this dataset can be utilized by Vision-Language Models (VLMs) [26] to interpret complex scenes. Given an image as input, VLMs can provide a textual description of the scene, identifying if an accident has occurred, is actively unfolding, or if there are signs of an imminent collision. This capability enhances the dataset's potential for developing sophisticated tools for accident detection and scene analysis in real-world traffic environments.

**Contributions.** To summarize, our main contributions are:

- We present the TUMTraf-Accid3nD dataset, a dataset curated specifically for rare and hazardous traffic incidents, providing a groundbreaking resource for accident-specific 3D perception research.
- TUMTraf-Accid3nD features 2,634,233 labeled 3D boxes, instance masks, and 2D box annotations with track IDs of real accidents across various lighting and weather conditions. This makes it the largest dataset focused on real-world accident scenarios captured in 3D.
- We introduce an annotation method that enables highly accurate and detailed labeling of accidents and near-miss events. The annotation process includes precise 2D and 3D bounding boxes, instance segmentation, and tracking of all traffic participants in the scene.
- We propose a framework to detect accidents in real-time and in different weather and lighting conditions.
- We support eight tasks with our dataset: object recognition, 2D object detection, accident classification, instance segmentation, 3D object detection, tracking, accident anticipation, and trajectory prediction.
- We provide baseline results for the first six tasks and open source our dataset, detection framework, and dev kit.

## 2. Problem Statement

High-quality datasets focused on accident scenarios are essential for advancing road safety. However, existing datasets rarely cover real accident cases in sufficient detail, making this an underrepresented but critical area of research. Most autonomous driving datasets focus on normal driving conditions to avoid the complexities of actual accident scenarios. This limits the potential for models to effectively learn and predict high-risk events. To improve safety in autonomous systems, there is a pressing need for a specialized dataset that captures the complexity of accident scenarios in diverse conditions. Such a dataset would provide a robust foundation for developing algorithms capable of understanding, detecting, and ultimately helping to prevent accidents.

Table 1. **Overview of publicly available accident datasets.** We compare the TUMTraf-Accid3nD dataset with other available synthetic and real accident datasets based on the following criteria: year, type (synthetic or real), perspective, number of image frames (#Img), available point clouds (PCs), number of 2D bounding boxes (#2D BB), number of 2D instance masks (#Masks), number of 3D bounding boxes (#3D BB), and track IDs (T). Entries with * indicate approximation based on average video frame counts reported in respective papers.

| Dataset | Year | Type | Perspect. | #Img | PCs | #2D BB | #Masks | #3D BB | T |
|---|---|---|---|---|---|---|---|---|---|
| ◦ VIENA² [28] | 2019 | synth | vehicle | **2,250,000** | × | × | × | × | × |
| ◦ GTACrash [29] | 2019 | synth | vehicle | 751,610 | × | × | × | × | × |
| ◦ MP-RAD [30] | 2023 | synth | roadside | 366,600 | × | × | × | × | × |
| ◦ RiskBench [31] | 2024 | synth | vehicle | N/A | × | × | × | × | √ |
| ◦ DeepAccident (DA) Dataset [32] | 2024 | synth | V2V & V2I | 57,000 | √ | **285,000** | × | **285,000** | √ |
| ◦ Accident Image Analysis (AIAD) [33] | 2018 | real | variable | 10,480 | × | × | × | × | × |
| ◦ Car Accident Det. & Pred. (CADP) [34] | 2018 | real | roadside | 75,030* | × | × | × | × | × |
| ◦ Causality in Traffic Accident (CTA) [35] | 2020 | real | vehicle | 342,495* | × | × | × | × | × |
| ◦ Car Crash Dataset (CCD) [36] | 2020 | real | vehicle | 75,000 | × | × | × | × | × |
| ◦ Acc. Det. CCTV Footage (CCTVF) [37] | 2020 | real | roadside | 990 | × | × | × | × | × |
| ◦ Argus Dataset [38] | 2021 | real | roadside | 120,000 | × | × | × | × | × |
| ◦ YoutubeCrash [39] | 2021 | real | vehicle | 7,720 | × | × | × | × | × |
| ◦ IITH Road Accident Dataset [40] | 2022 | real | roadside | 127,138 | × | × | × | × | × |
| ◦ TAD [41] | 2022 | real | roadside | 24,810 | × | × | × | × | × |
| ◦ Accident Detection Model (ADM) [42] | 2023 | real | roadside | 3,250 | × | × | × | × | × |
| ◦ MM-AU Dataset [43] | 2024 | real | vehicle | **2,190,000** | × | 2,233,683 | 2,233,683 | × | × |
| ◦ TUMTraf-Accid3nD Dataset (**ours**) | 2025 | real | roadside | 111,656 | √ | **2,634,233** | **2,634,233** | **2,634,233** | √ |

## 3. Related Work

Existing accident detection methods have never been tested on real roadside traffic data of an ITS test stretch. Real accident datasets are rare and do not contain enough data to train deep learning models [27], as shown in Table 1.

### 3.1. Accident Datasets

Several accident datasets [28, 29, 32, 33, 34, 36, 37, 38, 39, 40, 41, 42, 43, 44, 45] have recently been released. However, most are limited to 2D annotations, lacking realistic 3D labeling needed for comprehensive accident analysis. Although some 3D datasets exist, they are synthetic and not representative of real-world conditions.

The DeepAccident [32] dataset contains 691 synthetic accident scenarios in the CARLA simulator. These accidents were generated based on crash reports published by the National Highway Traffic Safety Administration (NHTSA). The dataset contains labeled data from four vehicles and one roadside infrastructure camera. One limitation of this work is that all accidents are generated in a simulation environment and do not represent realistic crash scenes. Hence, the sim-to-real gap must be addressed to improve the generalization capabilities of the perception models.

MM-AU [43] is a dataset for multi-modal accident understanding in videos. It contains 11,727 ego-view accident videos and 2.23 million 2D object boxes but lacks 3D box annotations, instance masks, and track IDs. Our dataset addresses this by providing high-quality 3D labeled real-world accident data in high-speed highway scenarios.

### 3.2. Accident Detection

Accident detection aims to identify the time and location of accidents within video frames, which is challenging due to rapid object motions, visual occlusions, and viewpoint changes caused by camera movements during crashes [46]. Early detection methods [47] primarily rely on frame-level appearance changes or simple motion cues to identify accidents, but these approaches struggled in complex environments. More recent methods emphasize spatiotemporal modeling, capturing motion consistency and scene evolution across frame sequences to improve robustness [43, 48]. Supervised methods train deep networks to classify accident vs. non-accident frames, while unsupervised techniques, such as DoTA [49], detect abnormal motion patterns by predicting future trajectories and flagging deviations between predicted and actual motions. Despite these advancements, most existing approaches rely on monocular 2D video data [50], losing important depth and spatial cues. This limits their ability to estimate object distances and collision risk accurately, reinforcing the need for real-world 3D accident datasets to enhance detection performance and generalizability.

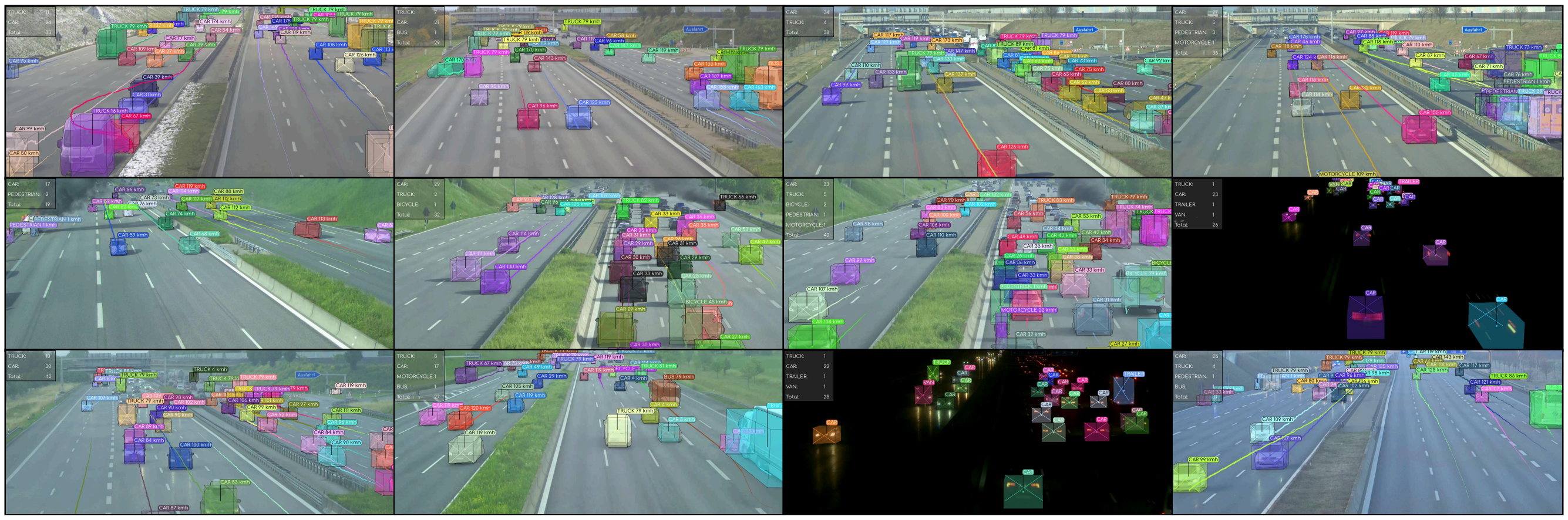

Figure 3. **Visualization of the labeled TUMTraf-Accid3nD dataset** with 3D box annotations, instance masks, track IDs, and trajectories. Accidents are recorded from roadside cameras on a test bed for autonomous driving. The dataset includes scenes with collisions and overturning vehicles, with some vehicles even catching fire.

### 3.3. Accident Anticipation

Accident anticipation aims to predict accidents before they occur, providing early warnings based on evolving scene dynamics. These methods typically process sequential video frames and estimate future accident likelihood using recurrent or temporal convolutional networks. To improve interpretability, models like DSA-RNN [51] introduce attention mechanisms that dynamically focus on critical objects and regions contributing to accident risk. Later works further enhance these attention mechanisms by incorporating driver behavior, such as gaze direction or steering patterns, to capture human reactions to emerging hazards. Methods like DRIVE [52] explicitly model interactions between traffic participants using spatiotemporal graphs, improving anticipation through relational reasoning among vehicles, pedestrians, and infrastructure. However, existing approaches rely heavily on annotated accident time windows, which are expensive to obtain and prone to ambiguity. Moreover, the absence of large-scale, realistic 3D accident datasets hinders generalization across diverse driving scenarios, emphasizing the urgent need for diverse 3D benchmarks to enable robust and transferable anticipation systems.

## 4. The TUM Traffic Accid3nD Dataset

This section presents the TUM Traffic Accid3nD (TUMTraf-Accid3nD) dataset, a real-world dataset of rare, hazardous traffic incidents for accident-specific 3D perception tasks.

### 4.1. Sensor Suite

The roadside infrastructure sensor setup is designed to continuously monitor traffic flow and detect accident scenarios using a diverse suite of sensors. Positioned at strategic locations, the sensor suite used for the dataset includes nine sensors: four high-definition cameras, four radars, and one LiDAR. These sensors are mounted on two sensor stations, such as the one shown in Fig. 4, to capture real-time traffic data from multiple perspectives. Each sensor type contributes uniquely to a multi-modal data fusion framework. The cameras provide high-resolution visual information, enabling object detection, tracking, and scene interpretation. Radars add velocity and range information, capturing the motion dynamics of vehicles, even in challenging weather conditions like fog or heavy rain. The LiDAR offers precise 3D spatial data, creating a detailed map of object positions and surroundings. Together, these sensors deliver a rich, fused dataset that enhances the accuracy of accident detection and analysis by combining visual, motion, and depth information. Through careful calibration, the data from each sensor is aligned in a common coordinate system to ensure a reliable multi-modal fusion. This setup enables scene understanding in various lighting and weather conditions and supports advanced applications like trajectory prediction, cooperative perception, and accident anticipation.

### 4.2. Data Collection Process

The data collection process includes several key stages: data recording, extraction, and anonymization. We capture high-quality data and ensure privacy and usability.

**Data Recording.** Data is continuously captured from roadside cameras, LiDAR, and radar sensors to monitor traffic flow and detect accident events during the day and night. Each sensor records data at 25 FPS to provide details for high-speed accident analysis. The recorded data is com-pressed and saved in *rosbag* files on secure servers in real-time. Each sensor stream is time-synchronized, ensuring that data across cameras, LiDAR, and radar align accurately.

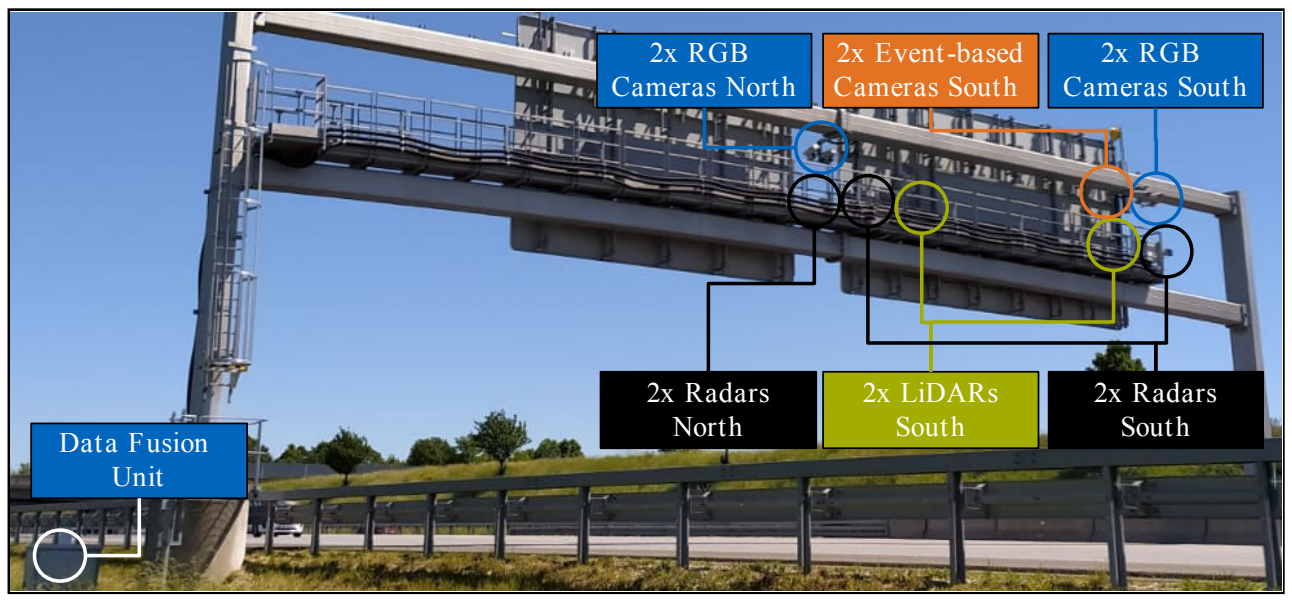

Figure 4. **Visualization of the sensor setup.** We show one of the two gantry bridges used to record the highway data with accidents.

**Data Selection.** Our data selection system uses a rule-based accident detection framework to prioritize capturing accident and near-miss events. By monitoring indicators like lane deviations, sudden speed changes, and unusual vehicle interactions, the system flags high-priority events and focuses on relevant accident scenarios. The selected data is then extracted into individual frames or point cloud scans.

**Data Anonymization.** To protect privacy, all data undergoes an anonymization process. License plates and personally identifiable information like faces are blurred. A YOLOv8 [53] network was trained on local license plates to detect them in real-time.

The dataset development kit, which includes pipelines for processing, loading, and managing data, enables researchers to efficiently access, preprocess, and work with the dataset.

### 4.3. Labeling Process

We use 3D BAT [54], an automatic 3D bounding box annotation toolbox, to annotate our dataset. It includes a detection and tracking step and a manual quality check to enhance accuracy. This tool automates the labeling process by first applying a custom 2D object detector [14] (based on YOLOv7 [55]) and a 3D object detector (MonoDet3D [14]). The objects are tracked with the PolyMOT [56] tracker and fused using a late fusion approach [14]. Each labeled traffic participant contains a 2D box, a corresponding instance mask, and 3D box information (position, dimensions, and rotations) with a track ID and speed value. To ensure high labeling quality, we manually inspect the generated annotations in the labeling tool, adjust them accordingly, and finally export them in the OpenLABEL [57] standard.

### 4.4. Coverage and Scenario Diversity

Our dataset features 111,945 labeled camera and LiDAR frames with over 2.6 million 2D and 3D box annotations, each accompanied by track IDs, trajectory data, and classification across six different instance types: cars, trucks, buses, pedestrians, motorcycles, and bicycles. It captures a diverse range of accident scenarios, including high-speed lane changes leading to rear-end collisions, vehicle overturning upon impact, and vehicles catching fire. Additionally, it includes instances involving emergency response vehicles and near-miss events. Example cases are illustrated in Figure 1. The dataset serves as an essential resource for developing and validating AI-based detection, tracking, data fusion, and trajectory prediction algorithms, as well as understanding the occurrence and after-effects of naturally occurring high-speed crash incidents and other accidents on highways.

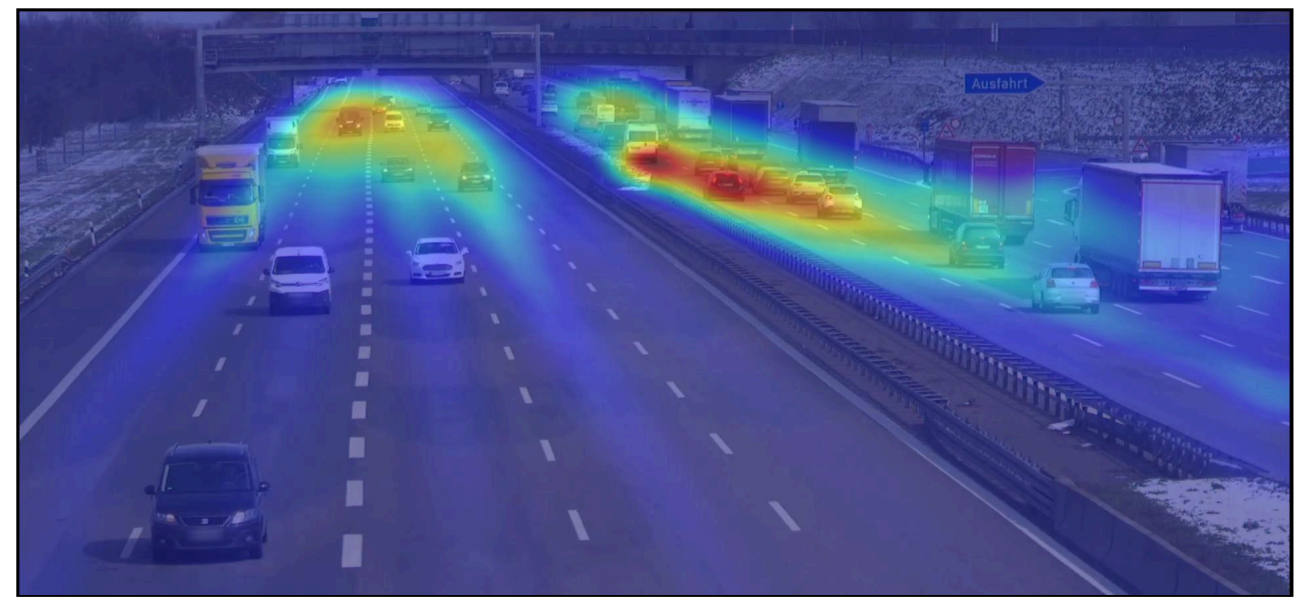

Figure 5. **Heatmap visualization of traffic participant locations.** The left lane on the highway towards the north direction indicates a high traffic volume.

### 4.5. Annotation Schema

Our annotation schema is based on the OpenLABEL [57] standard, structured to store 3D labels with detailed tracking information, event data, and attributes that document accident scenarios. Each 3D label includes tracking IDs to uniquely identify objects across frames and to capture the trajectories of road users over time. Each annotation holds attributes that mark the sensor ID, track history, speed, and number of 3D points inside the 3D box. This schema ensures an accurate, multi-dimensional representation of accidents and enables detailed analysis of incident sequences and individual participant behavior throughout each event.

### 4.6. Dataset Statistics

Our dataset provides a large collection of annotated accident events and diverse object types across 111,945 camera and LiDAR frames, with over 2.6 million 3D box annotations with track IDs. This large-scale dataset includes six object classes: cars, trucks, buses, motorcycles, bicycles, and pedestrians. The occurrence of labeled accident events allows for the evaluation of detection models under realistic, high-speed highway scenarios.

Figs. 5, 6, 7, 8, 9, 10, and 11 provide a detailed statistical overview: Fig. 5 illustrates a heatmap of traffic participants to highlight busy lanes with a high traffic density. The left lane contains a high traffic volume because of a van that has a breakdown on that lane. Figure 6 presents the distribution of object classes. Cars and trucks are the most commonly

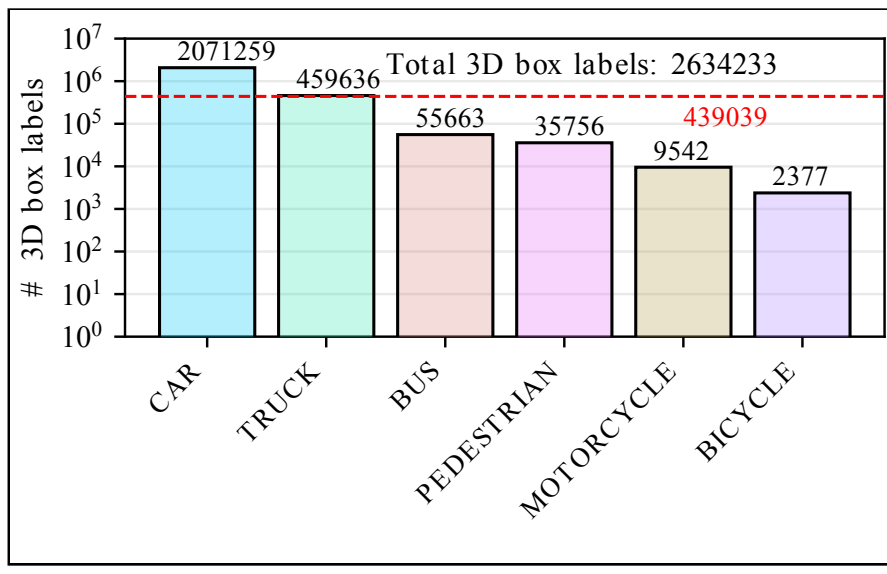

Figure 6. Distribution of object classes.

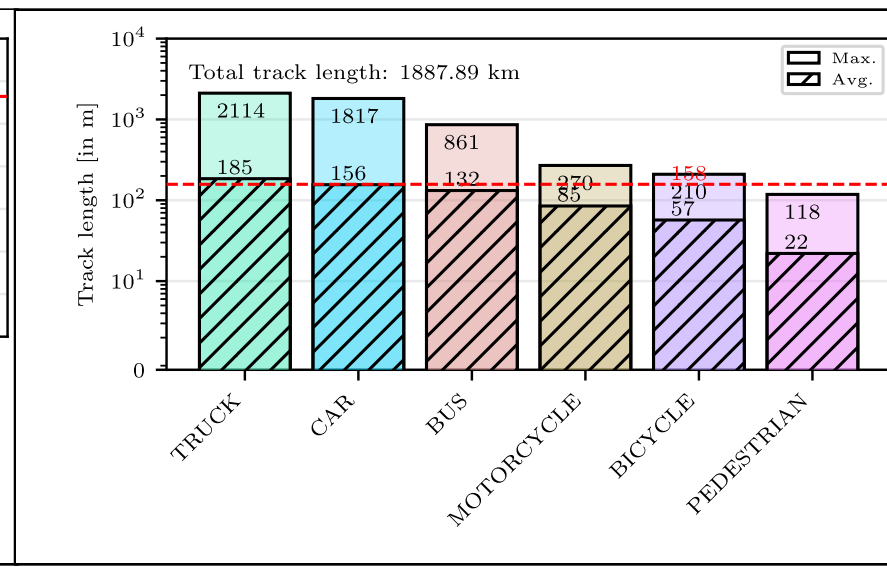

Figure 7. Average and max. track lengths of all labeled object classes.

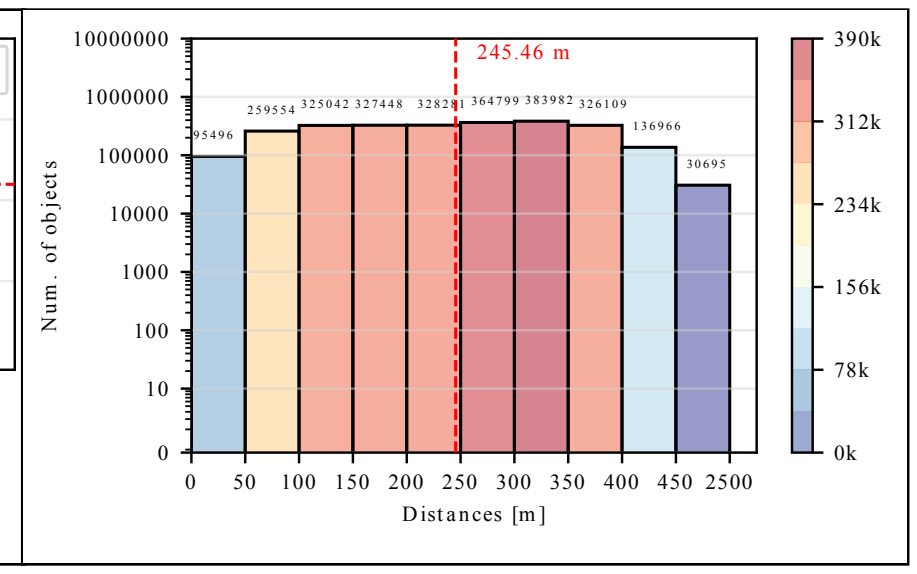

Figure 8. Histogram of labeling distances.

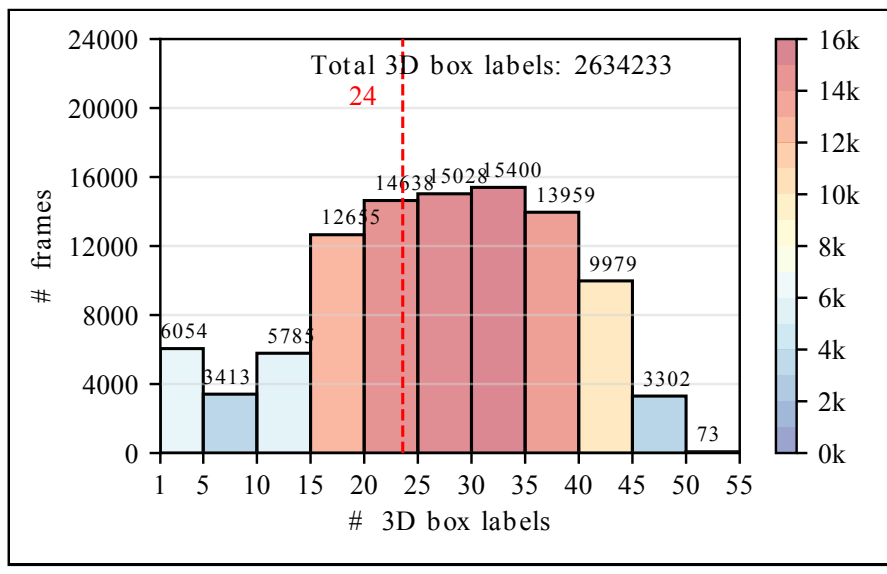

Figure 9. Histogram of the number of 3D box labels.

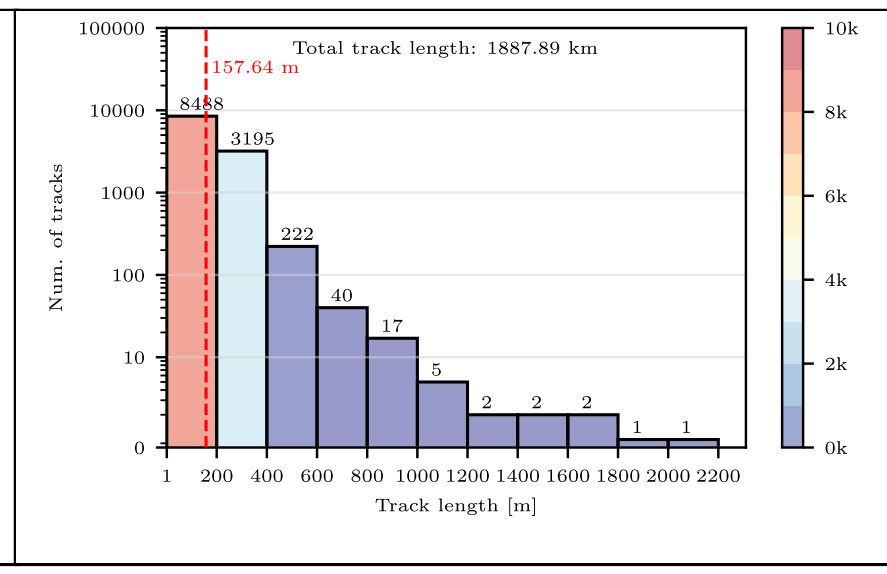

Figure 10. Histogram of the track lengths.

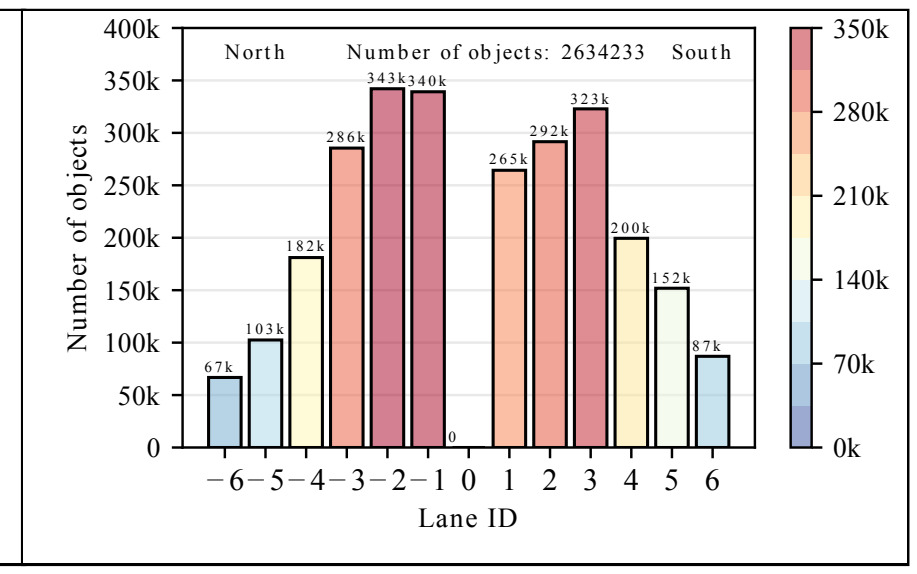

Figure 11. Lane distribution of all labeled objects on the highway.

labeled objects in the dataset. Fig. 7 shows average and maximum track lengths. Track lengths vary, with most up to 200 meters (see Fig. 10) and one truck reaching a 2,114-meter-long track. The total accumulated track length of all labeled objects is 1887.89 km on the highway. Fig. 8 shows the distribution of the labeling distances. On average, objects are 245 meters away from the sensor. Fig. 9 illustrated the histogram of the number of 3D box labels. Most frames contain 15–45 labeled objects, with up to 52 objects in a single frame and an average of 24 objects per image frame. The lane distribution analysis in Fig. 11 shows that the majority of labeled objects are in the southbound lanes, particularly in lane −2. The highway has 12 lanes in total, six lanes heading north and six lanes heading south.

### 4.7. Comparative Analysis

Unlike most 3D perception datasets, which typically focus on general traffic and non-accident scenarios, our dataset is dedicated to real-world accidents and near-miss events. This emphasis on critical situations allows for detailed modeling and analysis of incidents, setting it apart from popular 3D perception datasets like KITTI [58, 59], nuScenes [60], and Waymo Open Dataset [61]. While these datasets capture diverse traffic objects and scenes, they include no labeled accident events, limiting their use in the development of safety-focused AVs. Our dataset's unique strengths include its rich 3D annotations of accidents, detailed tracking information, and a variety of accident types such as high-speed collisions, multi-vehicle pile-ups, and emergency responses. The inclusion of labeled multi-sensor data enables robust multi-modal data fusion for scene understanding. We provide long tracks, capturing trajectories across multiple frames and supporting applications in trajectory prediction and risk assessment, which are essential for accident anticipation. This dataset offers a great resource for advancing research in safety-critical situations, filling a gap in the field with data curated specifically for accident detection and prevention.

## 5. Methodology for Accident Detection

Our framework consists of two modules, the detection and tracking module and the accident detection module. Fig. 2 visualizes the process from detecting objects in raw images to finally detecting accident events.

### 5.1. 3D Object Detection and Tracking

Our framework uses 3D perception techniques and multi-sensor data fusion to create a real-time digital twin of the traffic. Starting with raw camera images, the framework first performs 3D object detection using MonoDet3D [14] to identify and localize vehicles in three dimensions. Following detection, PolyMOT tracking [56] is applied to maintain continuity across frames, while sensor data fusion combines inputs from four roadside cameras and four radars, leading to generated trajectories and track histories.

Table 2. Evaluation of two object detection methods on our test set.

| Model | mAP@[.5] | mAP@[.5:.95] | mIoU | FPS |
|---|---|---|---|---|
| ◦ YOLOv7x-seg | 78.50 | 53.40 | 85.72 | 22 |
| ◦ YOLOv8x-seg | **80.10** +1.60 | **58.50** +5.10 | **88.50** +2.78 | **62** |

Table 3. Maneuver detection rules for the rule-based approach.

| | |
|---|---|
| $v_i \geq \frac{15 \text{ km/h}}{3.6}$ | Vel. of vehicle (i) |
| $v_i > v_{\text{lead},i}$ | $v_{\text{lead},i}$: Vel. of lead vehicle |
| $v_i \geq v_j \quad \forall i < j \leq N$ | $d_{\text{lead},i}$: Distance to lead vehicle |
| $d_{\text{lead},i} \geq d_{\text{thresh}}$ | $d_{\text{thresh}}$: Predef. distance threshold |
| $d_{\text{lead},i} < \left(\frac{v_i - v_{\text{lead},i}}{30}\right)^2$ | $\text{TTC}_{\text{lead},i}$: Time-to-Coll. lead veh. |
| $\text{TTC}_{\text{lead},i} \leq \text{TTC}_{\text{thresh}}$ | $\text{TTC}_{\text{thresh}}$: Time-to-Coll. threshold |

### 5.2. Accident Detection Pipeline

Our accident detection pipeline combines rule-based and learning-based methods to detect and classify accidents in real-time using roadside infrastructure data. This dual-module approach uses raw images and the trajectories of the digital twins that were produced with 3D perception, tracking, and fusion modules to find accident events.

**Rule-Based Accident Detection**. This module analyzes tracked object trajectories, which are generated by the MonoDet3D and PolyMOT frameworks, as outlined in the previous section. It extracts features like lane IDs, distance matrices, and object velocities, identifying potential accidents through predefined maneuver detection rules (see Table 3). If all six rules apply simultaneously, the corresponding traffic participant is classified as an accident event. For example, sudden changes in speed or trajectory angle, unsafe lane changes, and close proximity situations are flagged as potential incidents. These rule-based detections provide immediate alerts and enable rapid incident response.

**Learning-Based Accident Detection.** To enhance accuracy, the pipeline also includes a learning-based YOLOv8 object detector trained on our custom accident dataset. When the rule-based module signals a potential accident, the learning-based detector validates the event by analyzing the camera feed. YOLOv8 is fine-tuned to detect accident-specific cues such as collisions, overturned vehicles, and vehicle fires, producing classifications with associated confidence scores, object locations, and velocities. Detected incidents must appear in at least three consecutive frames with a confidence score above 0.8 to minimize false positives. Additionally, results from all available cameras in a scene are fused to enhance robustness. The output of this integrated accident detection pipeline includes a real-time accident classification for each detected vehicle, along with its location and other critical metadata. This pipeline, illustrated in Fig. 2, supports both training data preparation and live accident monitoring on highways.

Table 4. Evaluation of the PolyMOT 3D tracker on our test set.

| Model | FN | FP | MT | PT | ML | IDS | FRAG | HOTA | MOTA | MOTP |
|---|---|---|---|---|---|---|---|---|---|---|
| PolyMOT | 1313 | 657 | 20 | 58 | 50 | 19 | 50 | 0.45 | 0.18 | 1.63 |

Table 5. 3D detection results of MonoDet3D on our test set.

| Model | 3D mAP@[.1] |
|---|---|
| ◦ MonoDet3D + YOLOv7 (baseline) | 15.20 |
| ◦ MonoDet3D + YOLOv7 + PolyMOT | 16.23 +1.03 |
| ◦ MonoDet3D + YOLOv8 | 17.77 +2.57 |
| ◦ MonoDet3D + YOLOv8 + PolyMOT | **18.24** +3.04 |

## 6. Experimental Results

First, we outline the benchmark tasks and establish baseline performance metrics for 3D object detection, 3D tracking, and 3D accident detection on our dataset. Baseline performance metrics include 3D mean average precision ($\text{mAP}_{\text{3D}}$) for 3D detection, Multi-Object Tracking Accuracy (MOTA) for tracking, and F1 score values for accident classification.

### 6.1. Object Detection and Tracking Performance

Moreover, we evaluate two 2D detection and segmentation models on our test set using an input size of $1280^2$ px and TensorRT acceleration on an NVIDIA RTX 3090 GPU (see Table 2). The YOLOv8x-seg performs best in all metrics. We further evaluate the PolyMOT [56] 3D tracker on our test set and report the results in Table 4.

**Ablation Study.** We provide an ablation study on our test set for multiple 3D object detection baselines of MonoDet3D [14] in combination with different YOLO models and trackers. In Table 5 we can see that MonoDet3D performs best when employing the YOLOv8 model and PolyMOT tracker.

### 6.2. Accident Classification Evaluation

We recorded camera images and the fused perception results for 128 days, stored them in *rosbag* files, and processed these recordings. The automatic accident analysis was executed on 12,290 15-minute videos. Figure 12 shows the quantitative evaluation results. In total, 831,969 unique vehicles were identified (5,547 per day). 26.08% of vehicles were driving faster than the allowed speed limit of 120 km/h in the south direction. We found that outbound (north) traffic is often driving faster than inbound (south). The maximum detected speed was 264 km/h in the north direction of the highway where no speed limit is set. We detected 3,748 (0.45%) standing vehicles in a driving lane, 138 standing vehicles in a shoulder lane, and 120 breakdown events. Qualitative results of the rule-based accident detection are shown in Figure 13.

**Classification evaluation.** We first evaluate the accuracy and runtime performance of both accident classification models on our TUMTraf-Accid3nD dataset (see Table 6). The rule-based approach was able to detect 120 breakdown events. Four were false positives due to inaccurate object detections. This leads to a precision rate of 96.67%. On the other hand, the learning-based accident detection approach achieves a precision rate of 75.00%, which is limited by the relatively small training dataset, which consists of 3,725 images and 2,216 labeled accident events.

Table 6. **Accident classification** results and runtime evaluation of the RBA and LBA approach.

| Approach | Accuracy ↑ | Runtime ↓ [s] | |
|---|---|---|---|
| | F1-Score ↑ | 2 cam. | 4 cam. |
| Rule-based Approach (RBA) | 0.667 | **0.086** | **0.127** |
| Learning-based Approach (LBA) | **0.889** | 4.072 | 7.995 |

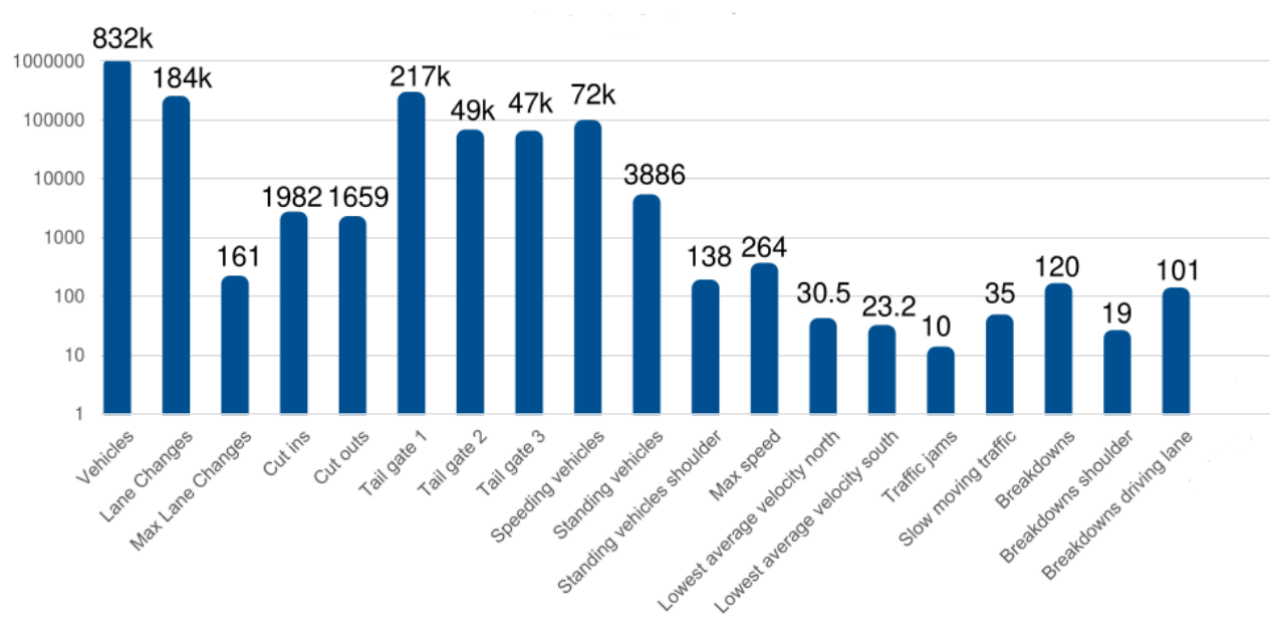

Figure 12. Quantitative results of the RBA accident detection module over a 128-day monitoring period. The statistics include the detected vehicles, maneuvers, scenarios, and accident events.

**Runtime evaluation.** The rule-based approach (RBA) runs at 95.05 FPS (10.41 ms per frame) on an NVIDIA RTX 3090 GPU. The total runtime for a 15-minute *rosbag* file with 22,500 ROS messages recorded at 25 FPS is 234.25 seconds. This includes the extraction of the lane ID, the distance calculation between all vehicles, and the scenario classification. In Table 7 we compare our two approaches based on the runtime. On average the RBA approach is about 57 times faster than our learning-based approach (LBA).

## 7. Conclusion and Future Work

Traffic accidents remain a leading cause of death worldwide, and the ability to rapidly detect accidents via roadside infrastructure sensors holds the potential to accelerate emergency response times and save lives. In this work, we introduced the TUMTraf-Accid3nD dataset, a resource focused on accident detection, and evaluated two detection methods on real-world data. Our dataset, detection framework, and dev kit are available as open-source tools on our project website to encourage widespread research and collaboration across academia and industry. By establishing TUMTraf-Accid3nD as a benchmark for accident detection, we aim to foster advancements in accident anticipation and detection, ultimately supporting the *Vision Zero* goal of eliminating traffic fatalities by 2050.

Table 7. **Runtime comparison.** We compare our rule-based and learning-based accident detection based on the runtime.

| Sequence | Runtime ↓ [s] | |
|---|---|---|
| | Rule-based | Learning-based |
| Sequence S01, part I | 8.63 | 484.69 |
| Sequence S01, part II | 6.60 | 474.69 |
| Sequence S13, part I | 5.02 | 240.43 |
| Sequence S13, part II | 5.33 | 248.16 |
| Avg. (with 2 cameras) | 5.17 | 244.30 |
| Avg. (with 4 cameras) | 7.61 | 479.69 |
| **Average (overall)** | **6.39** | 361.99 |

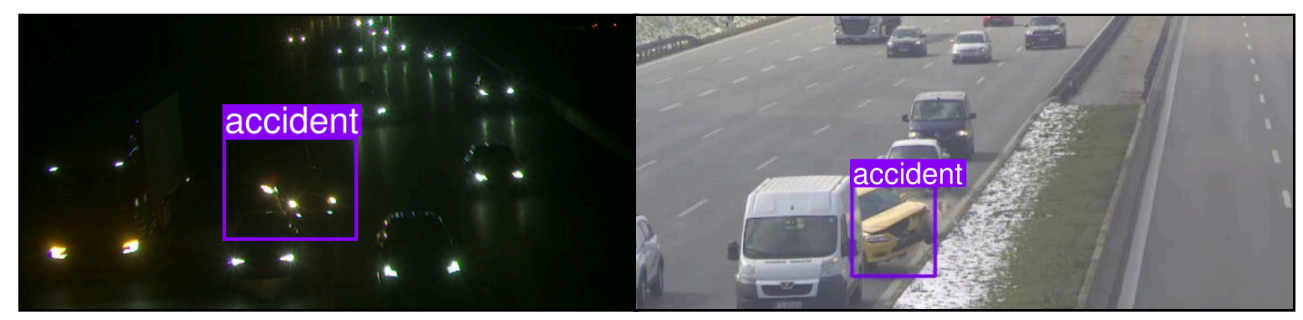

Figure 13. **Qualitative results of our accident detection framework on the TUMTraf-Accid3nD test set.** Left: The rule-based approach detected a rear-end collision. Right: The learning-based approach detected a car crash with a confidence score of 0.8.

**Future Directions.** We will expand our dataset to include more accident events under varied lighting and weather conditions, enhancing its applicability to complex real-world scenarios. We also plan to incorporate additional types of accidents in diverse environments, particularly urban areas where high-risk intersections often involve vulnerable road users (VRUs). By capturing events where VRUs are at risk due to behaviors like crossing against traffic signals or smartphone distraction, we aim to support additional safety applications. To increase the dataset's variety, we intend to expand the number of accidents recorded by five sensors to include data from 12 cameras on highways and another 12 in urban settings. Our design allows for integration with VLMs (see supplementary) to enable the interpretation and analysis of critical events, thereby supporting advancements in AV safety and intelligent traffic management. Finally, we will explore accident video diffusion methods [62] for generating realistic accident scenarios [63].

**Impact on AD Safety.** The TUMTraf-Accid3nD dataset and our detection framework will contribute to autonomous driving safety by enabling more accurate detection and prediction of accident scenarios. The availability of this dataset supports the development of robust algorithms aimed at reducing accident rates, improving system responses, and ultimately enhancing the safety of AVs on the road.

**Limitations.** Our RBA module currently focuses on rear-end collisions. We plan to address that by expanding the detection framework to capture additional accident types. Future improvements will enhance the adaptability to detect a broader range of accidents, further strengthening its value as a resource for autonomous safety research.

## 8. Acknowledgment

This research was supported by the Federal Ministry of Education and Research in Germany within the AUTOtech.agil project, Grant Number: 01IS22088U.

# Towards Vision Zero: The TUM Traffic Accid3nD Dataset
# Supplementary Material

accident-dataset.github.io

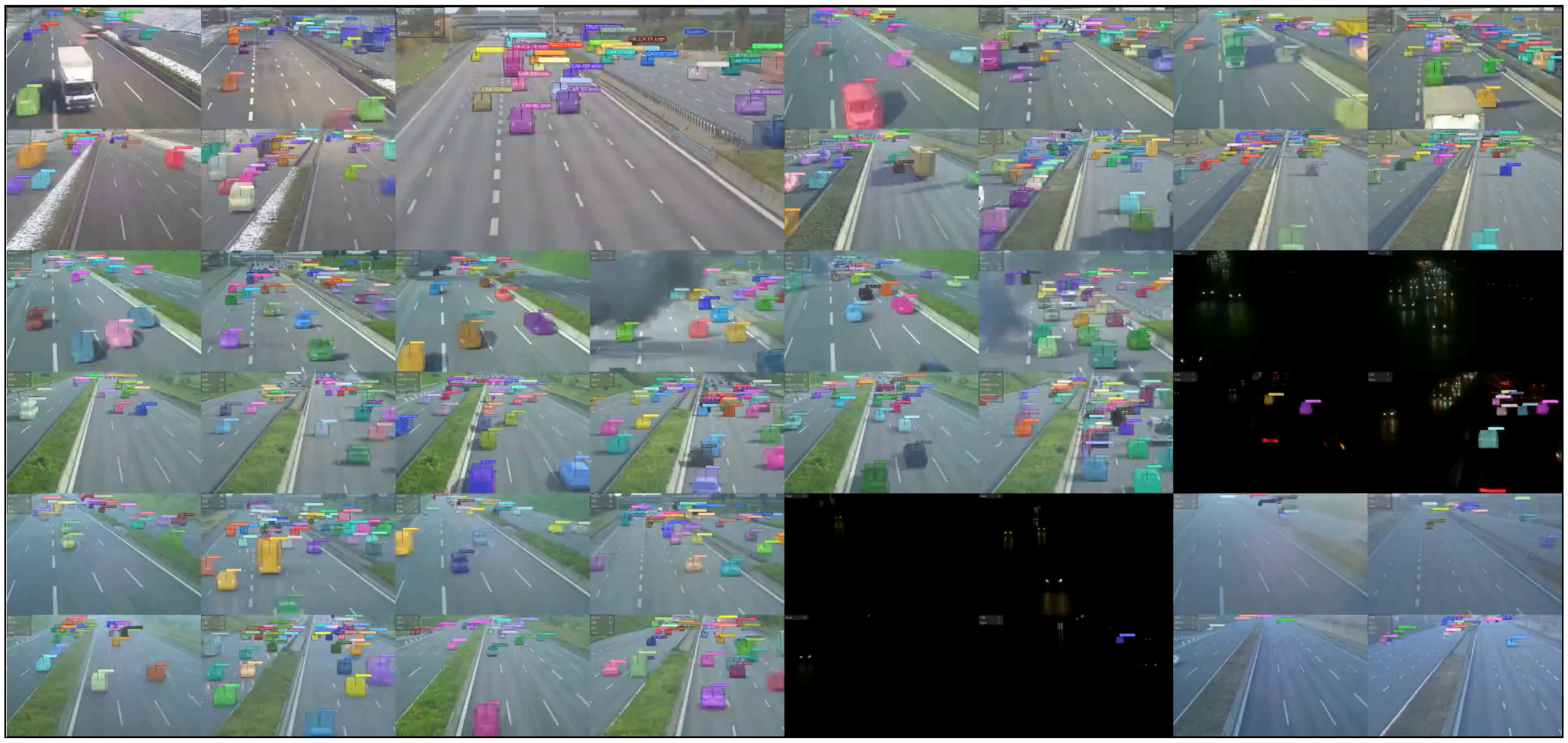

Figure 1.**Visualization labeled sequences within the TUMTraf-Accid3nD dataset.** It contains 12 sequences each recorded from four roadside cameras on a highway test bed. The dataset includes accident scenes during different lighting and weather conditions.

## Contents

## A. Additional Related Work

### A.1. VLMs for Accident Detection

Vision-language models (VLMs) [1, 2] are increasingly being explored for accident detection, as they enable a deeper understanding of complex scenes by linking visual data with language-based descriptions. They use natural language to describe objects, actions, and interactions in visual scenes. VLMs like AccidentGPT [3] can interpret and contextualize accident scenarios with enhanced detail. The integration of VLMs aids in recognizing accident-related events, such as sudden braking, near-misses, or unusual traffic patterns. With vision-language models, systems can also provide interpretable explanations for detected incidents, offering insights that support safer decision-making in autonomous and intelligent transportation systems. Related approaches were developed in [4, 5] to detect both accidents and general traffic anomalies.

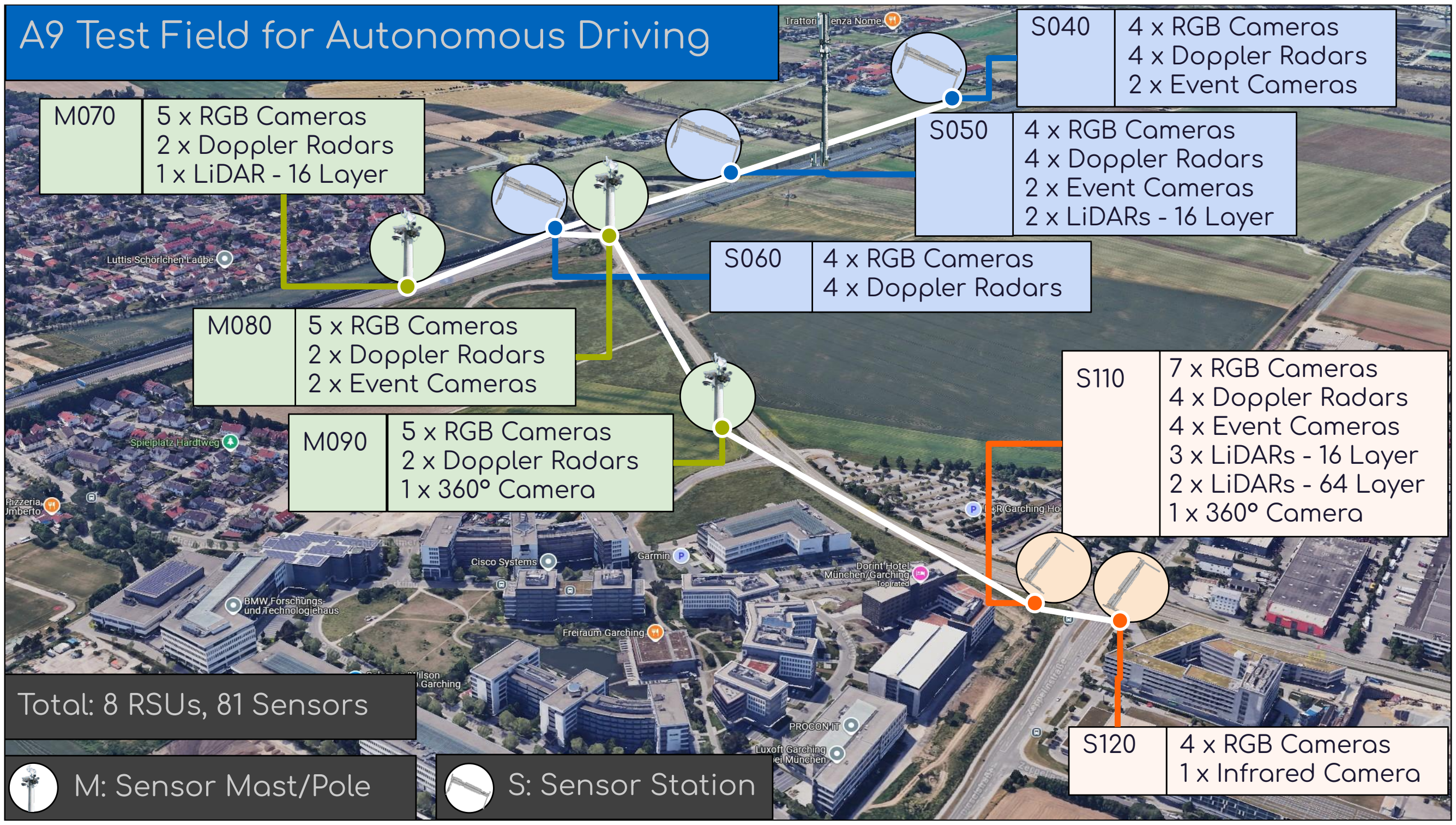

Figure 2. **Overview of the A9 Digital Test Field for Autonomous Driving.** The blue sensor stations (gantry bridges S040 and S050) on the highway were used to record the data with four roadside cameras, four radars, and one LiDAR.

## B. Extended Dataset Description

### B.1. Data Collection Setup

The A9 Digital Test Field for Autonomous Driving is located on the A9 highway in Munich, Germany. Figure 2 shows the 3 km long test field from a bird's-eye view. It contains eight measurement stations in total: three on the highway (marked in blue), three in a rural area (marked in green), and another two in an urban area (marked in orange). Each station includes a robust sensor suite with roadside cameras, radars, and LiDAR units. The infrastructure supports seamless tracking of objects across multiple sensor stations using sensor data fusion and tracking. All traffic participants are continuously monitored, and the anonymized recordings are stored on secure servers. Sensors were positioned to optimize coverage, enabling detailed observation of accident-prone areas such as merging lanes and intersections. The dataset was recorded at a busy section of the highway, containing 12 lanes including two exit lanes. On this highway, various scenarios can occur in diverse weather (sunny, cloudy, foggy, rainy) and lighting conditions (day, dusk, dawn, and night time). The specific accident types are described in Section B.3.

### B.2. Annotation Details

The dynamic highway environment posed significant challenges, particularly in ensuring accurate sensor calibration over a long time. Changes in harsh weather conditions, such as heavy rain, hail, or snow, have often required recalibrating the sensors to maintain an accurate data capture. Occlusion was another critical issue, as large trucks and dense traffic frequently obstructed smaller vehicles and pedestrians, complicating the annotation process. We apply novel tracking methods to track occluded objects. Additionally, accident scenarios often involved rolled-over vehicles, which made it difficult to annotate. We did not decide to annotate the roll and pitch angle of traffic participants to simplify the labeling guidelines. To ensure a high labeling quality, we adopted an iterative annotation process that incorporated feedback from domain experts.

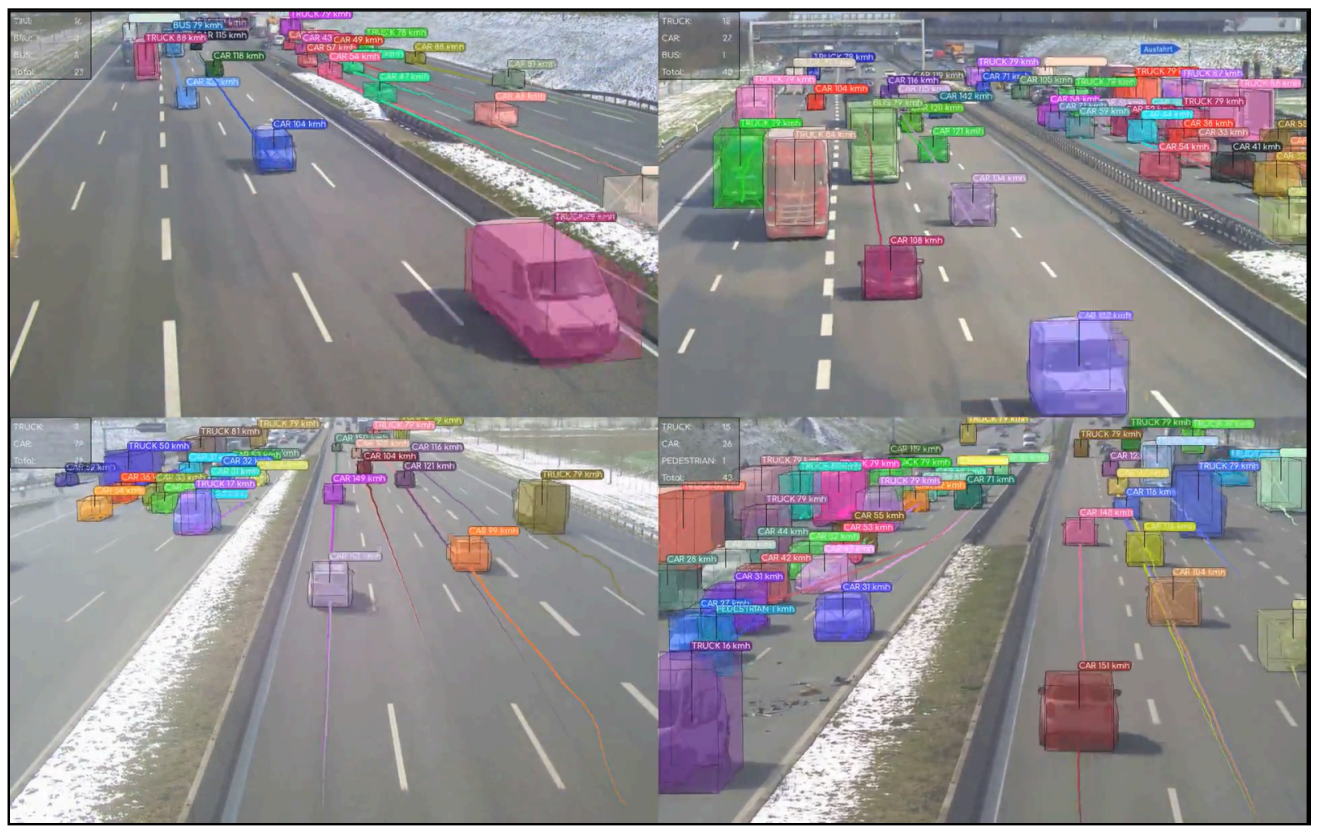

Figure 4. **Visualization of labeled sequence S01.** A vehicle is speeding at 180 km/h and crashes into a van that has a breakdown on the left lane.

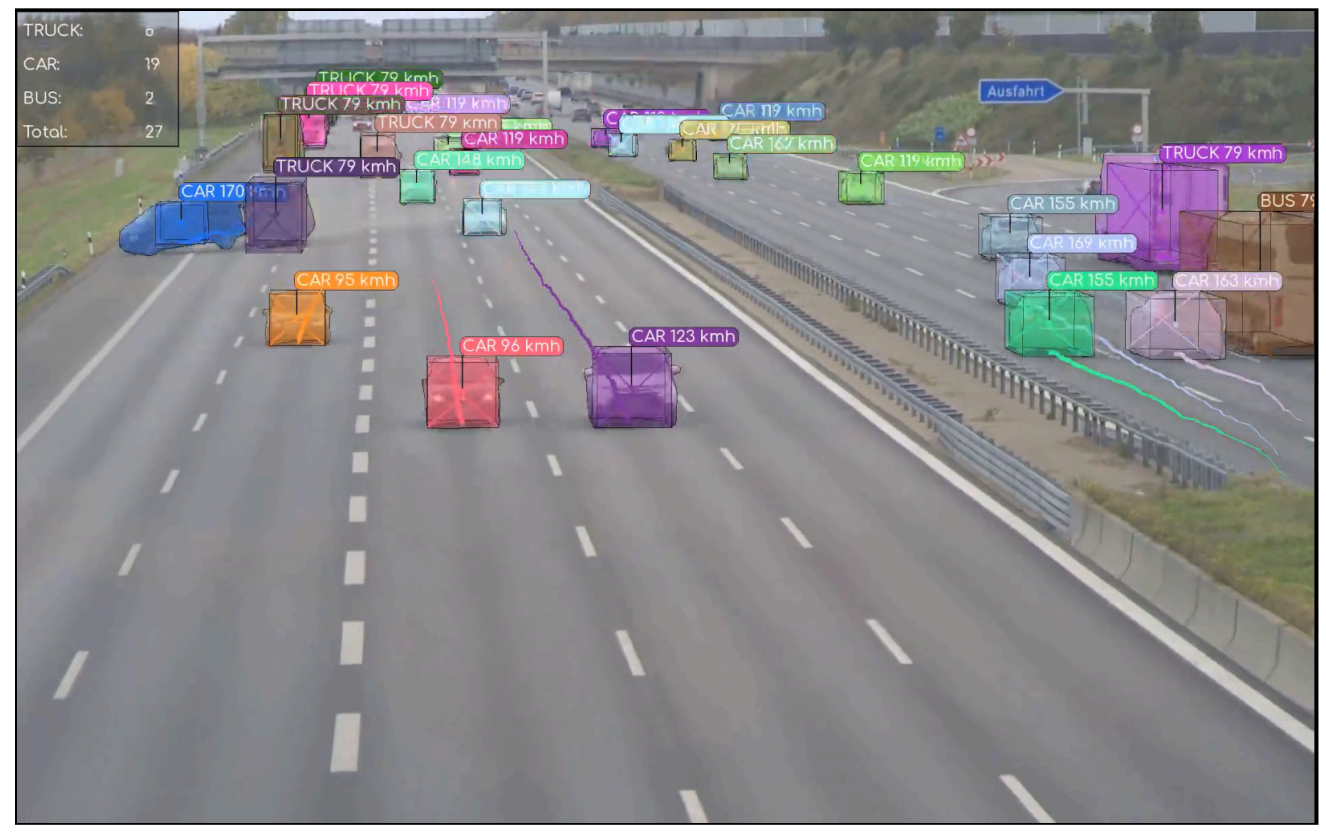

Figure 5. **Visualization of labeled sequence S02.** A blue van with a trailer is tipping over after a strong wind gust is hitting it.

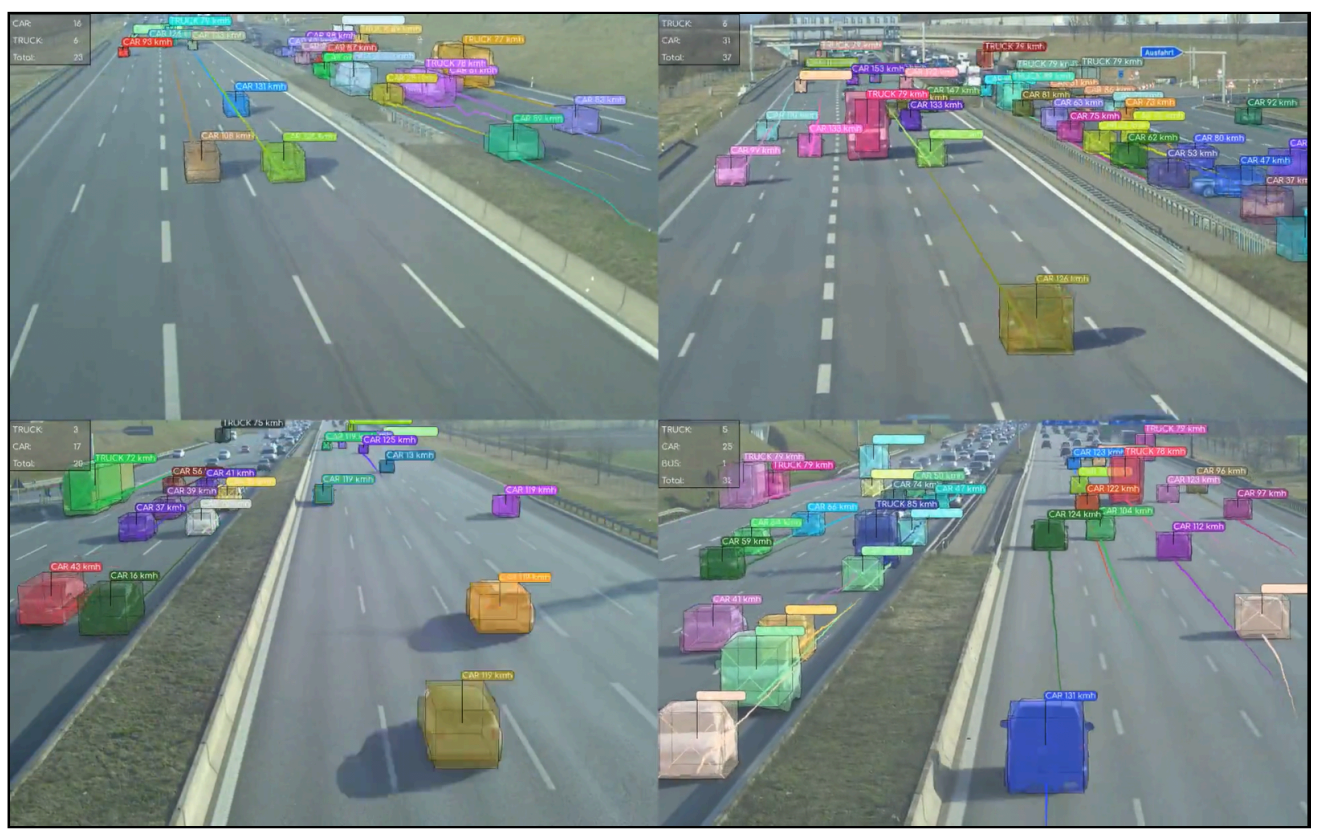

Figure 6. **Visualization of labeled sequence S03.** A vehicle is not maintaining safety distance to the leading vehicle. It changes the lane to the right and hits another two vehicles. One of them is full 360-degree spin.

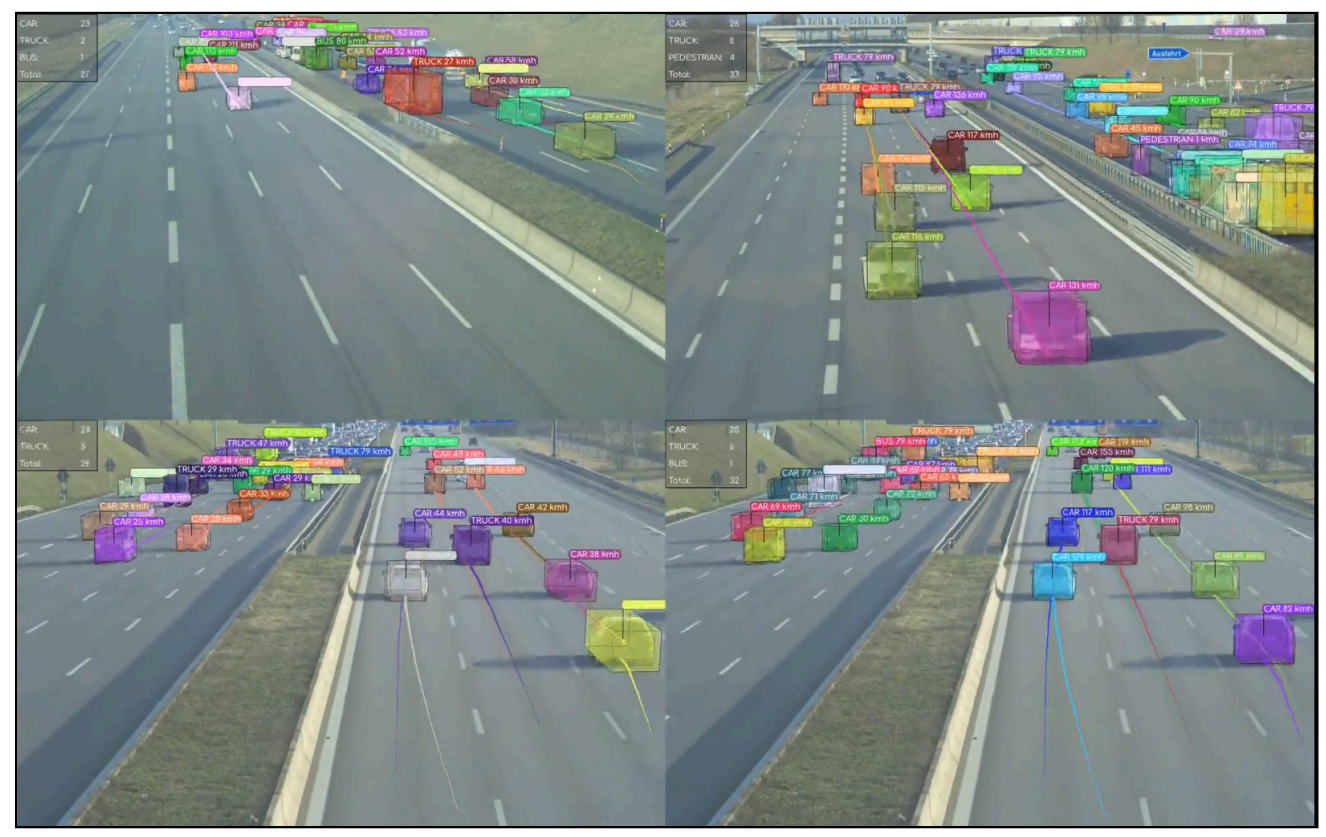

Figure 7. **Visualization of labeled sequence S04.** A post-accident scenario was labeled with arriving police, ambulance, and fire trucks. Some people get out of their cars to secure the site of an accident.

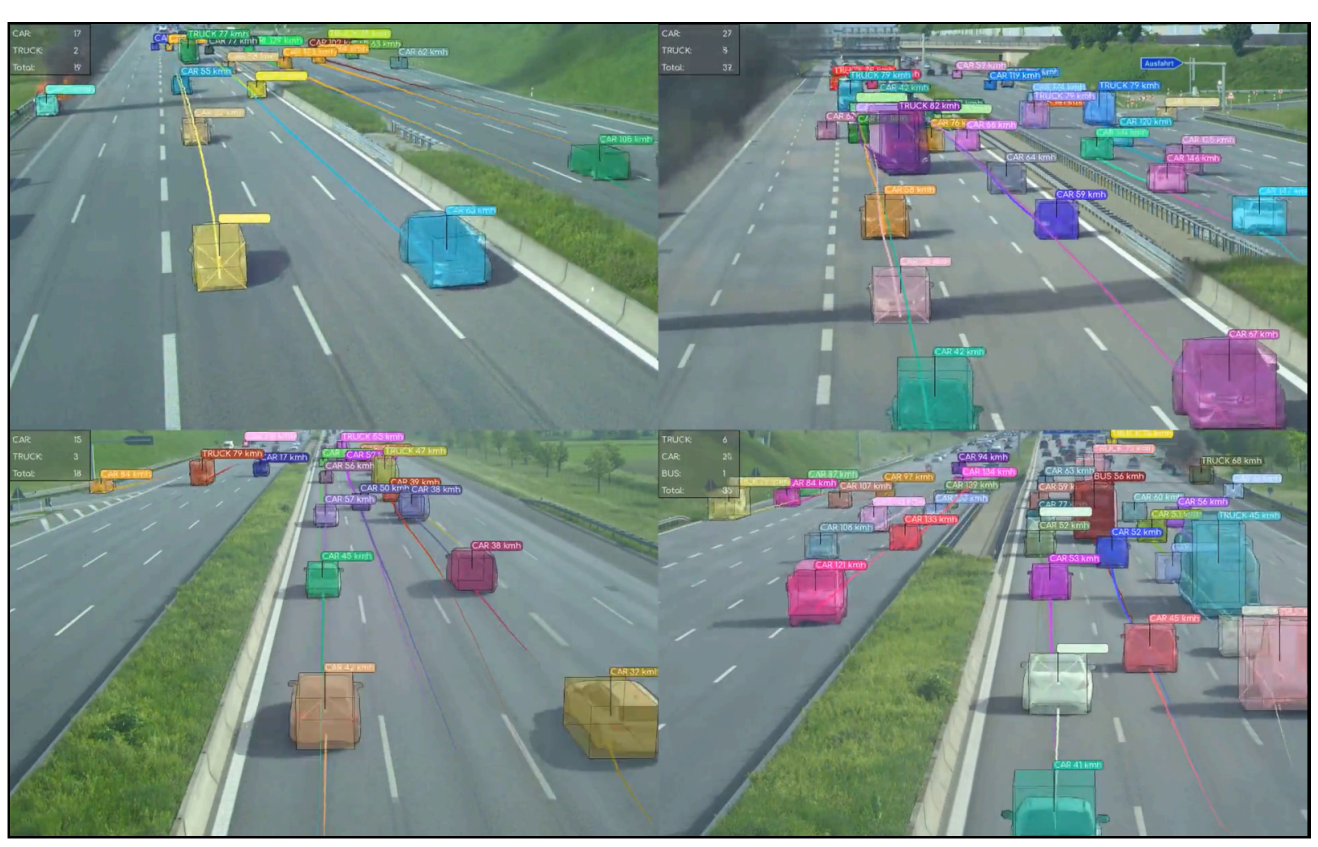

Figure 8. **Visualization of labeled sequence S05.** A van stops on the shoulder lane and starts to burn. The passengers are getting out of the van and secure their belongings. A man tries to extinguish the fire without any success.

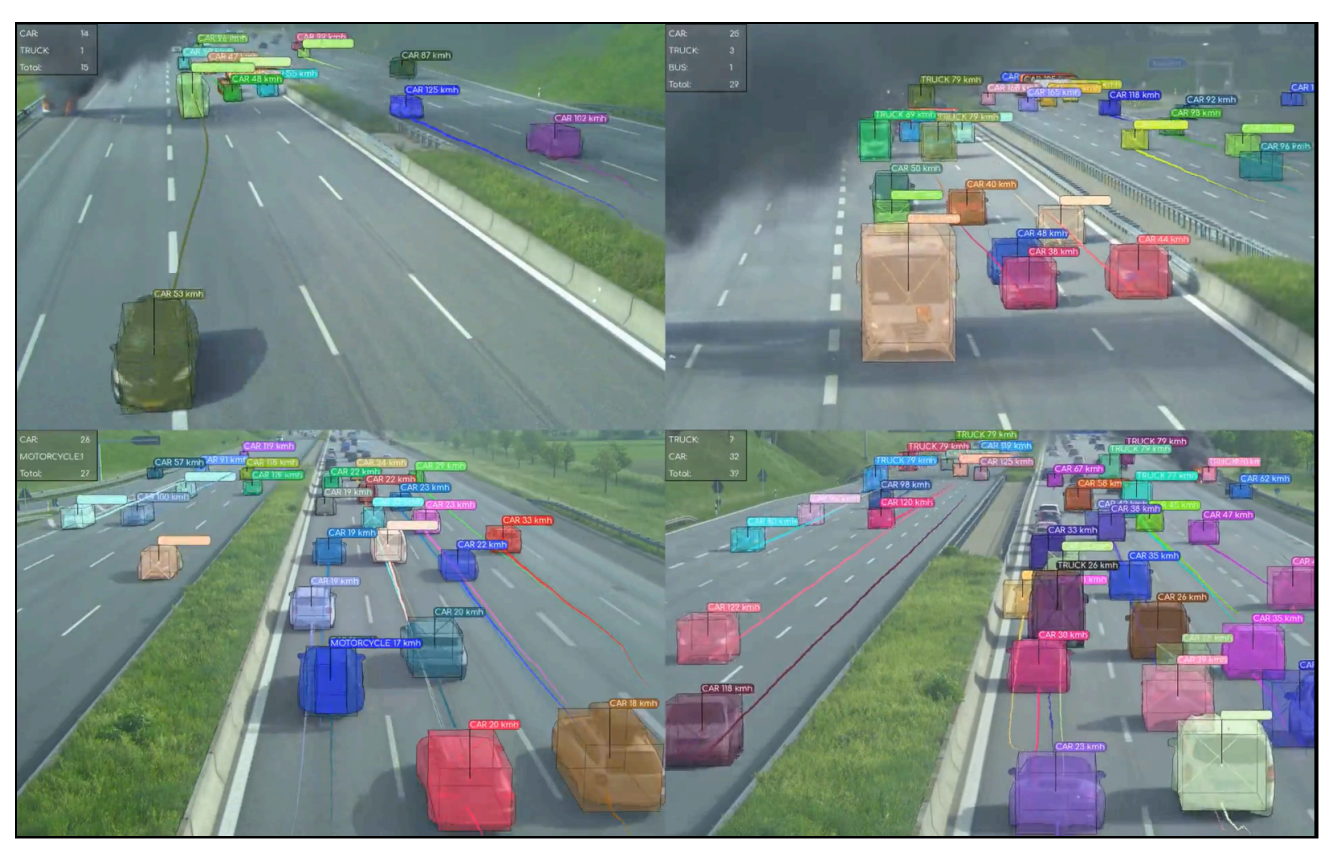

Figure 9. **Visualization of labeled sequence S06.** The van starts to burn heavily and produces a large smoke cloud that occludes the traffic participants in the roadside camera.

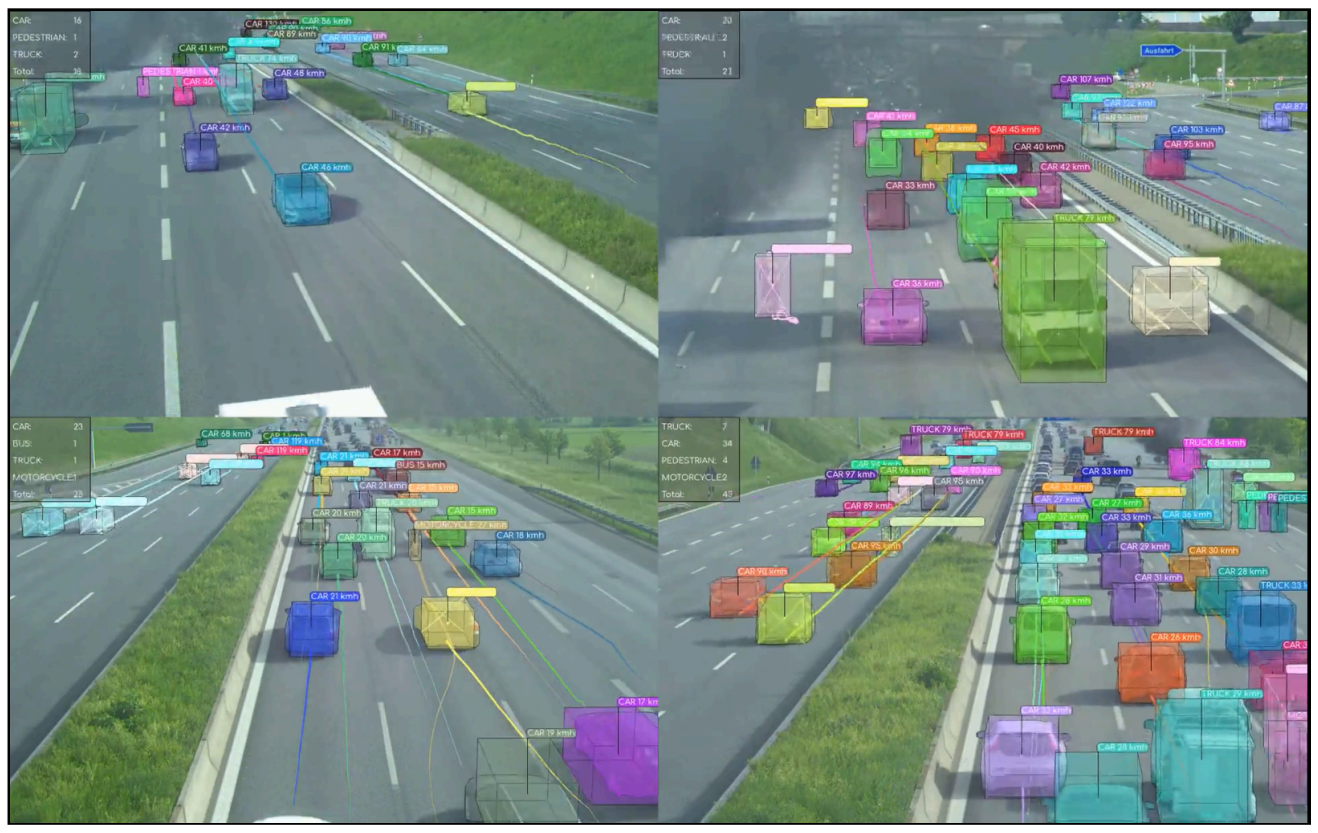

Figure 11. **Visualization of labeled sequence S07.** Emergency vehicles (police, ambulance, and fire trucks) arrive a van is burning for 15 minutes.

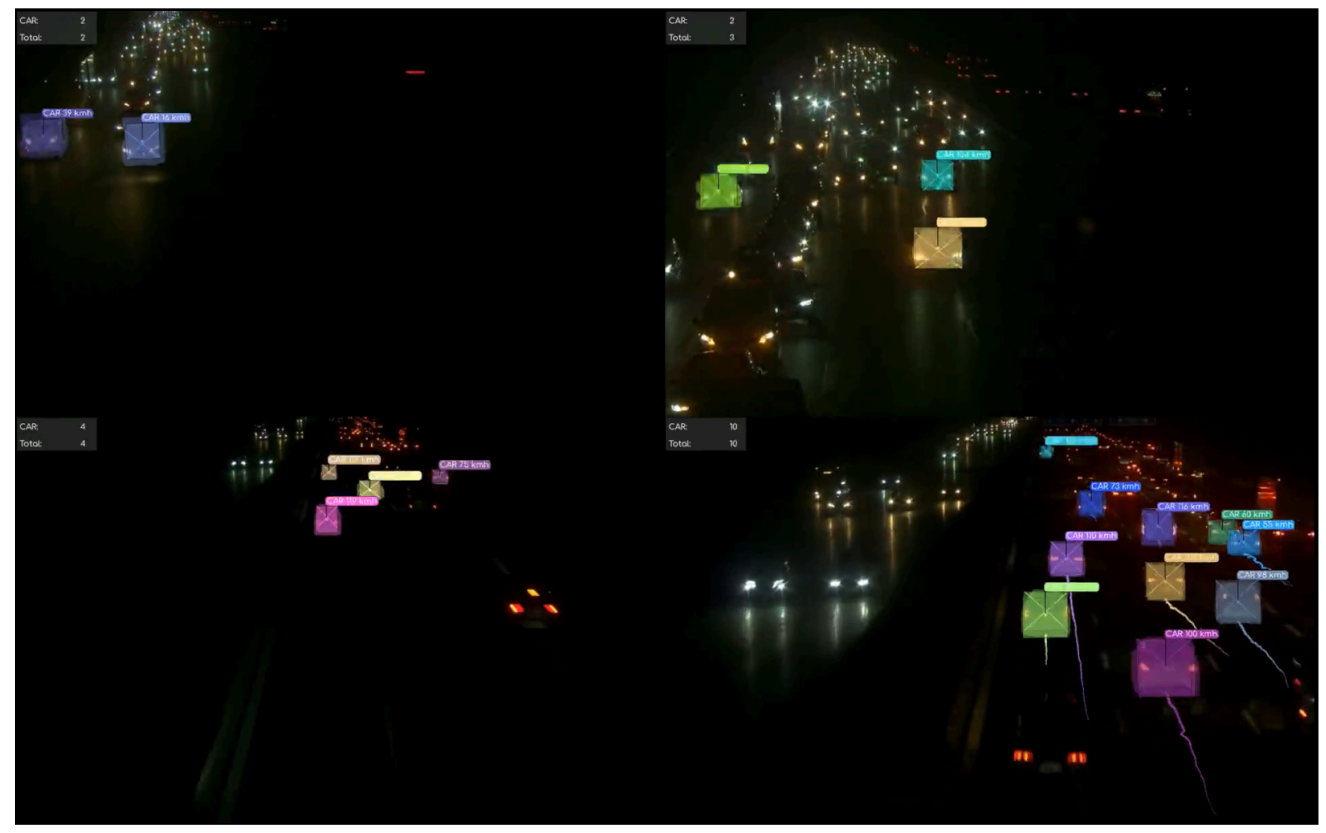

Figure 12. **Visualization of labeled sequence S08.** A traffic jam is forming on the highway. A vehicle changes lanes at night without turning on the indicator lights. Another vehicle is tail-gaiting that vehicle from the back.

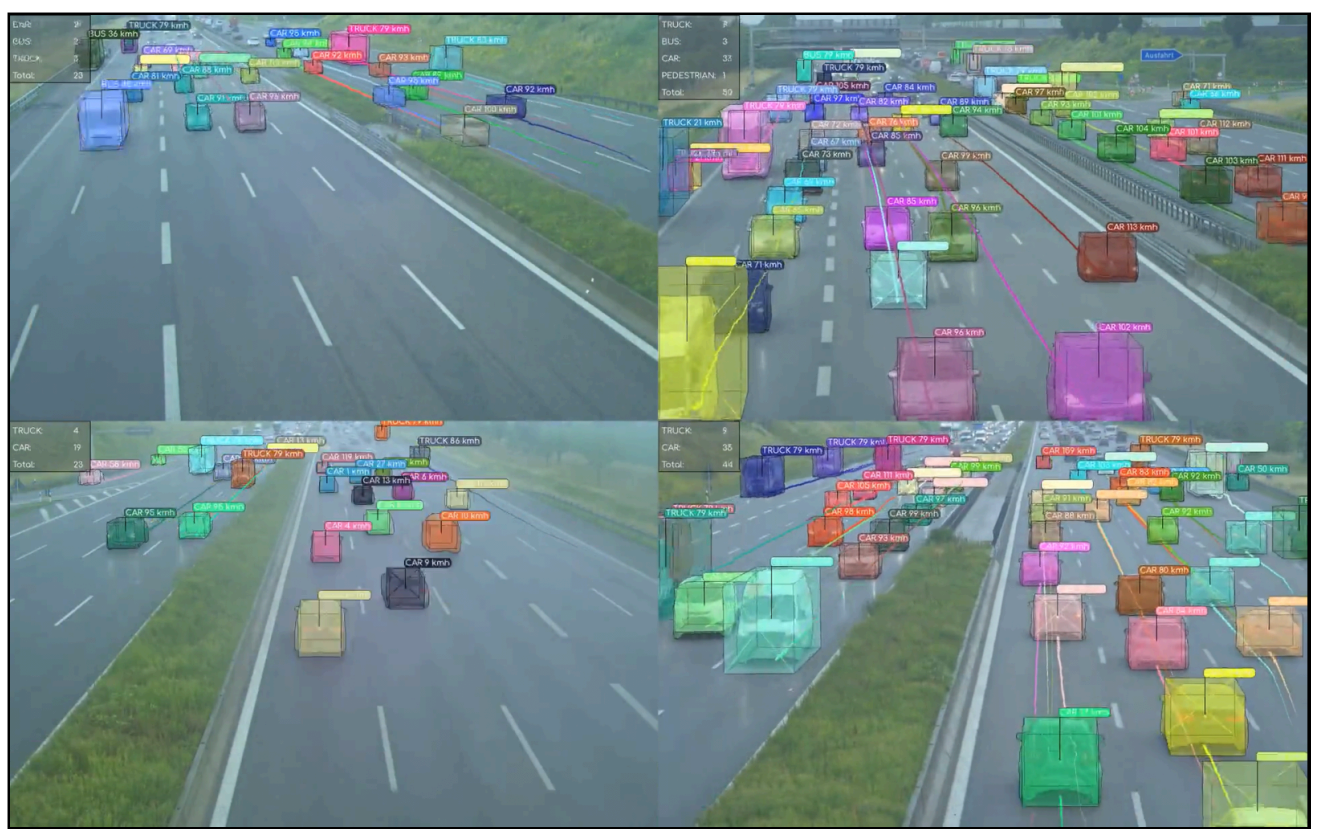

Figure 13. **Visualization of labeled sequence S09.** A truck is changing lanes and tailgates a car from the back. Both vehicles stop on the shoulder lane to inspect the collision.

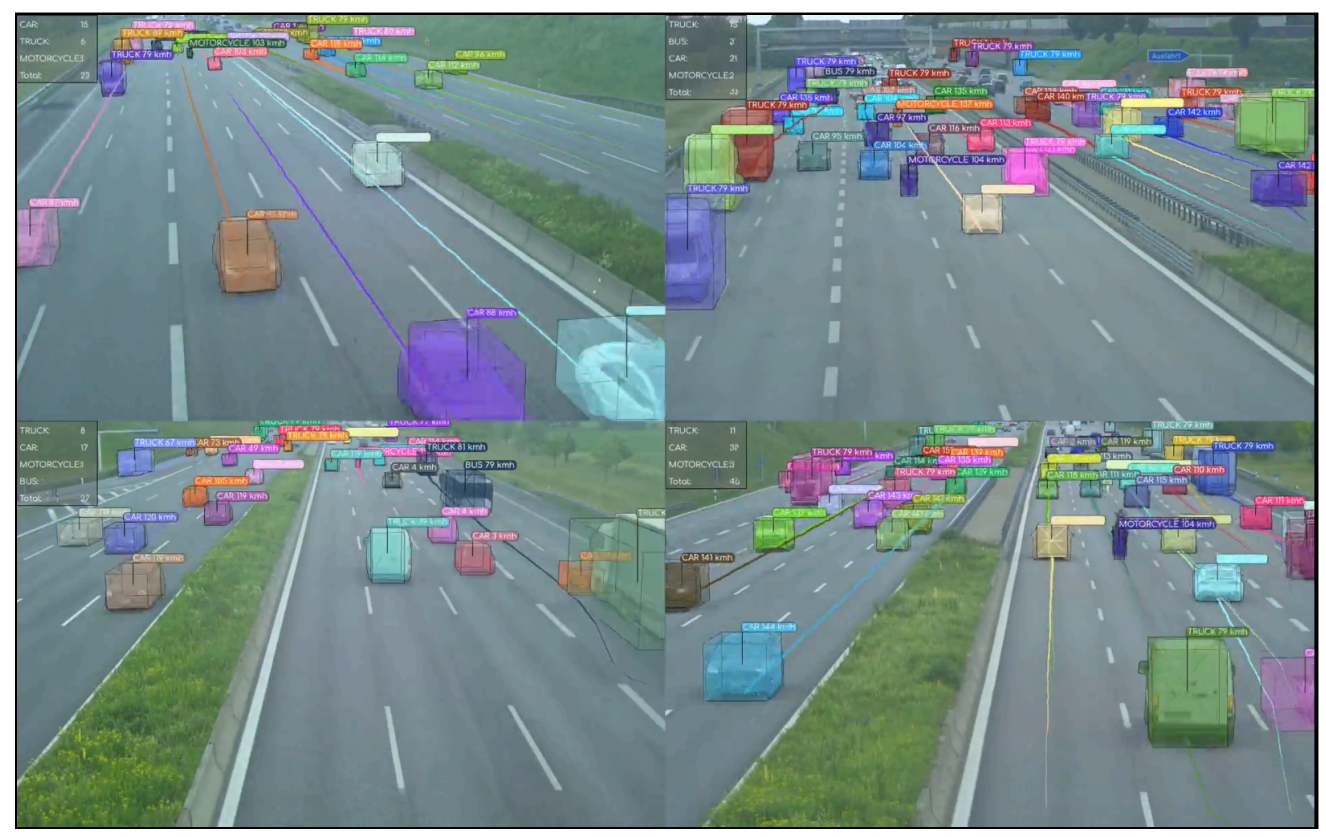

Figure 14. **Visualization of labeled sequence S10.** A truck is changing lanes and rams a passenger vehicle from the side. It is spinning 180-degree in one direction and then in the other.

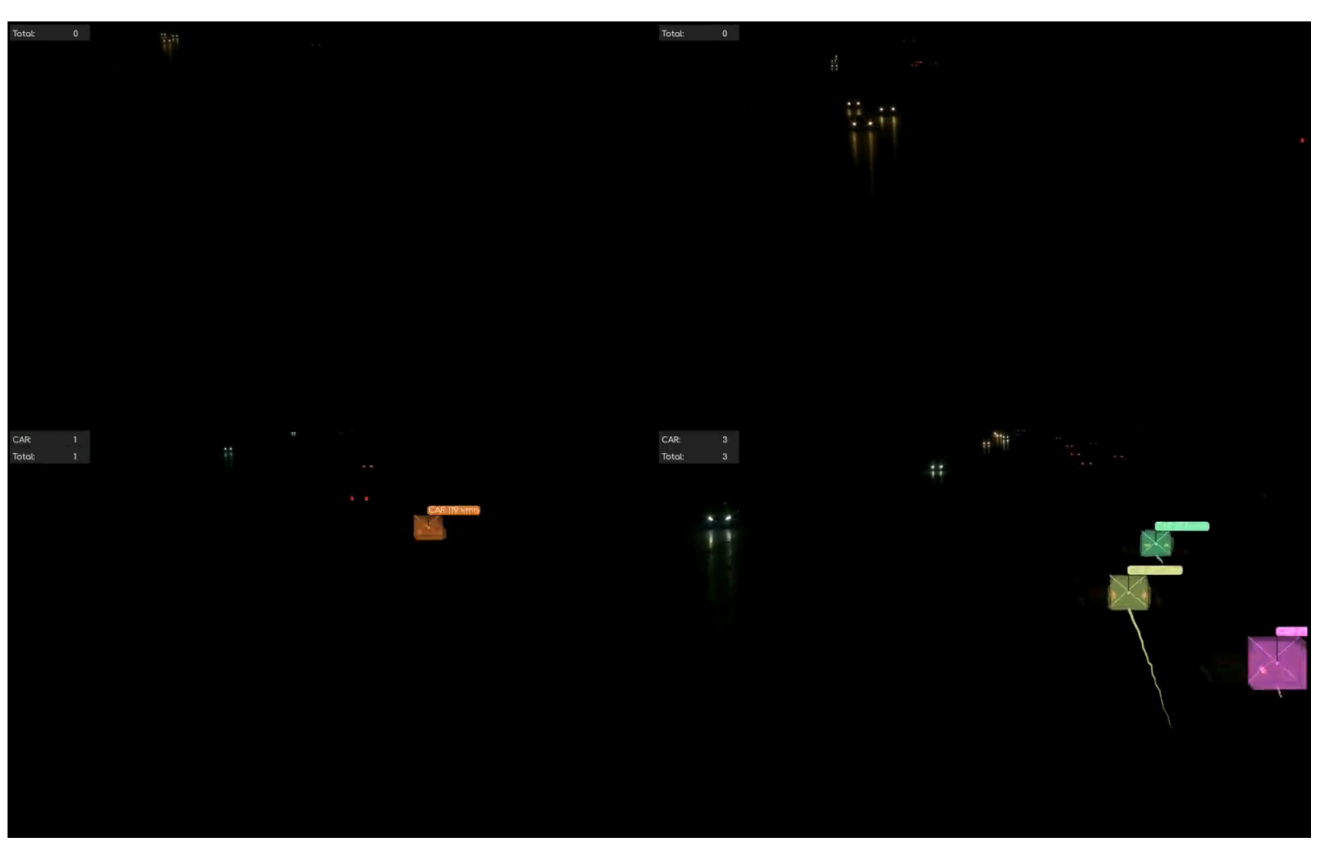

Figure 15. **Visualization of labeled sequence S11.** The driver of a pick-up truck falls asleep momentarily at night, changes three lanes, and flies over the guardrail. After landing behind the guardrail, it is rolling over three times.

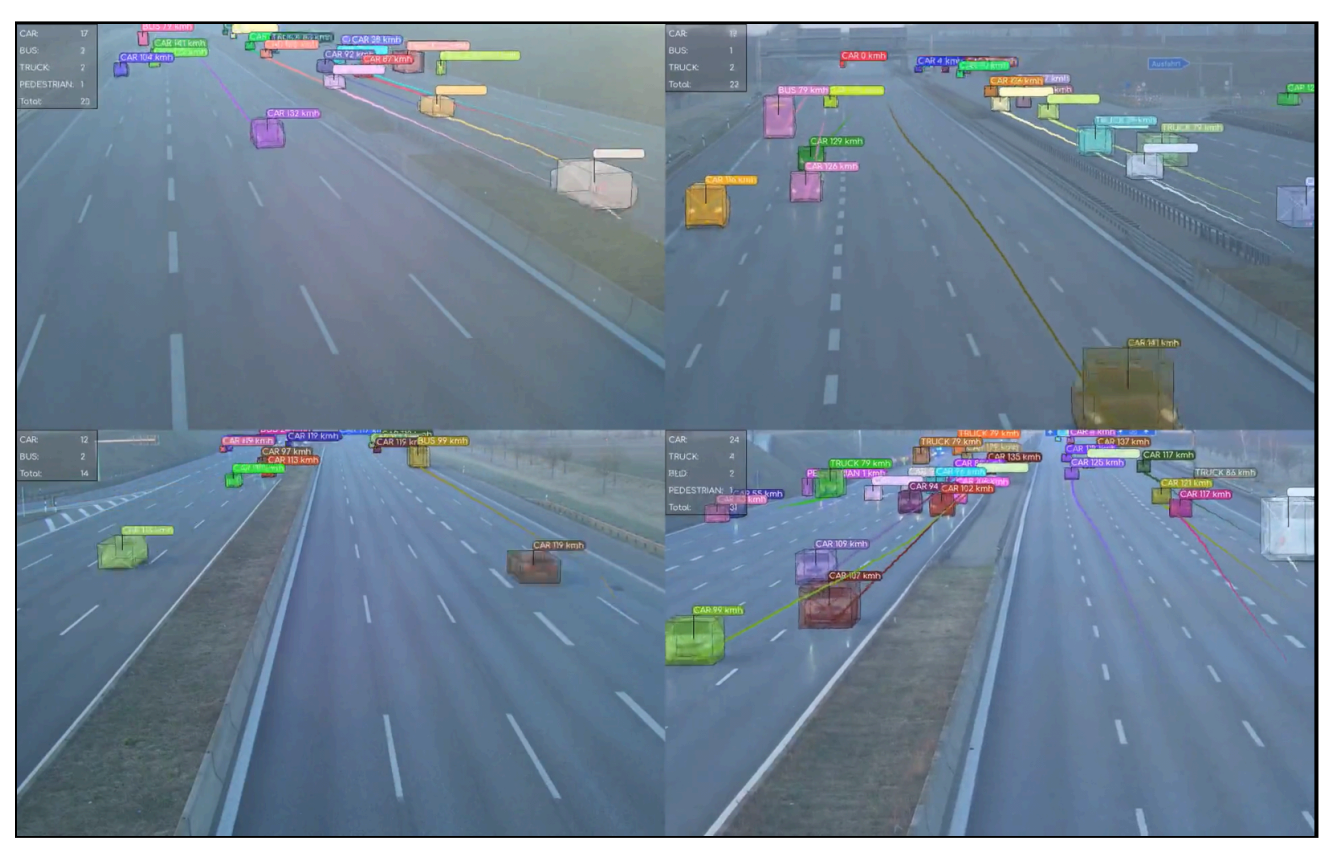

Figure 16. **Visualization of labeled sequence S12.** A vehicle is speeding and crashes into a van that has a breakdown on the left lane.

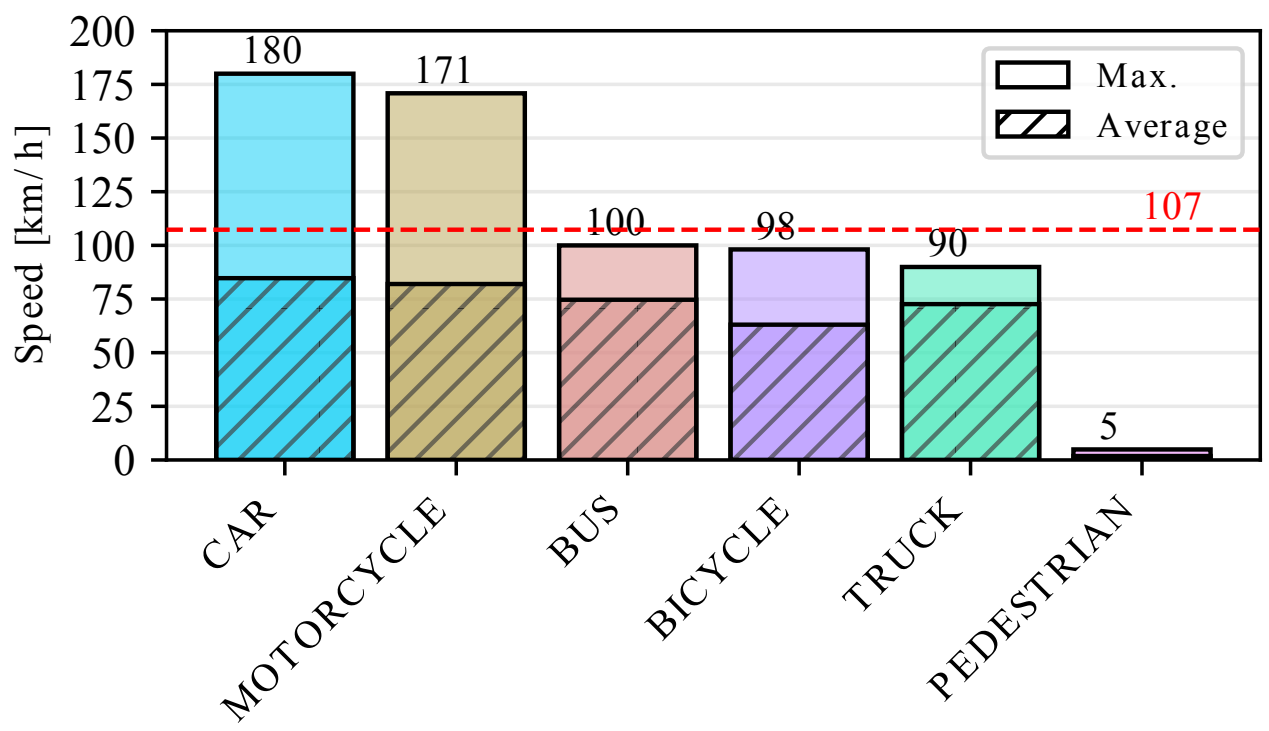

Figure 17. **Visualization of speed values for each labeled category.** We show the average and maximum speed values for all categories. The average speed is 107 km/h in the dataset.

### B.3. Quality Assurance

A multi-level quality assurance process was implemented to ensure annotation precision and consistency. Expert reviewers conducted manual inspections of labeled frames to identify misalignments or inaccuracies in categories, instance masks, and 3D bounding boxes. Automated checks were employed to validate trajectory continuity and to ensure logical consistency in object movements. For example, any abrupt jumps in tracked objects' paths were flagged for review. This hybrid approach combined the advantages of human expertise and automated validation, significantly improving the reliability of the dataset while reducing annotation errors.

### B.4. Detailed Dataset Visualization

Das dataset is visualized in Figs. 4, 5, 6, 7, 8, 9, 11, 12, 13, 14, 15, and 16. All 12 sequences recorded from four roadside cameras are displayed. The annotated frames in the dataset include the object category, speed, color-coded 2D instance masks and 3D bounding boxes to denote the object track, trajectory lines to indicate movement paths, and the total number of traffic participants in the current scene. These visualizations provide a view of interactions within a scene, such as the buildup to an accident or evasive maneuvers by other vehicles. These visualizations of traffic participants, coupled with speed values, make it easier to analyze and interpret complex traffic scenarios. Moreover, the speed and velocity of all traffic participants were calculated based on the 3D location and the time difference between the frames. Figure 17 visualized the average and max. speed values for each labeled category. The max. speed in the labeled dataset is 180 km/h (sequence S01). The overall maximum speed that was found in the recordings is 263 km/h in the north direction of the highway, where no speed limit is set.

## C. Comparative Dataset Analysis

### C.1. Quantitative Comparison

Our dataset is tailored for accident-centric research, addressing gaps left by popular datasets like KITTI [6, 7], nuScenes [8], Waymo Open [9], DAIR-V2X [10], and TUMTraf-V2X [11]. With over 111,000 labeled frames, it offers a large-scale dataset with a high density of safety-critical scenarios. Trajectories are notably longer with a maximum of 2,114 meters, allowing for the study of pre-accident behaviors such as abrupt lane changes or sudden decelerations. The dataset's multi-modal approach, combining data from four roadside cameras and one LiDAR, enhances its suitability for cooperative perception research. A comparative table (Table 1) illustrates these metrics, and emphasizes our dataset's strengths in annotation density, accident-specific focus, and environmental diversity.

### C.2. Unique Features

Our dataset stands out with its accident-focused labeling and unique features tailored for safety-critical analysis. It includes high-speed accident scenarios (up to 180 km/h) and extensive trajectory lengths, with a cumulative track length of 2,250 km and individual tracks exceeding 2 km. Captured on a busy 12-lane highway, frames contain up to 55 labeled objects that are labeled at 25 Hz with synchronized camera and LiDAR data. With over 2.6 million 3D boxes, the dataset precisely records 3D vehicle locations and enables realistic accident reconstructions in simulations. It can also be used to train models for detecting small and distant objects (200–400 m away) or to better detect objects that are occluded by large vehicles like trucks. Finally, our dataset includes diverse scenarios, such as night-time accidents, which allows robust model training.

### C.3. Prevalence of Accident Types

Accident scenarios are categorized into key types to support diverse research objectives. Rear-end and side collisions comprise 25% of accidents, often captured in stop-and-go traffic conditions. Breakdown events account for another 25%. They occur frequently on the highway, and vehicles stop on the shoulder lane to inspect the failure. Multi-vehicle pileups, where more than two vehicles were involved, also contained rear-end and side collisions. They occurred in 16.67% and are extensively annotated to capture their complex dynamics. In another two sequences (16.67%), the driver fell asleep. One sequence (8.3%) contains a scenario in which a heavy wind burst caused an accident. A trailer that was towed by a van tipped over, letting the vehicle tip over. The pie chart in Figure 18 illustrates the distribution of these accident types, offering a clear visualization of the dataset's focus areas and the breadth of scenarios included for safety analysis.

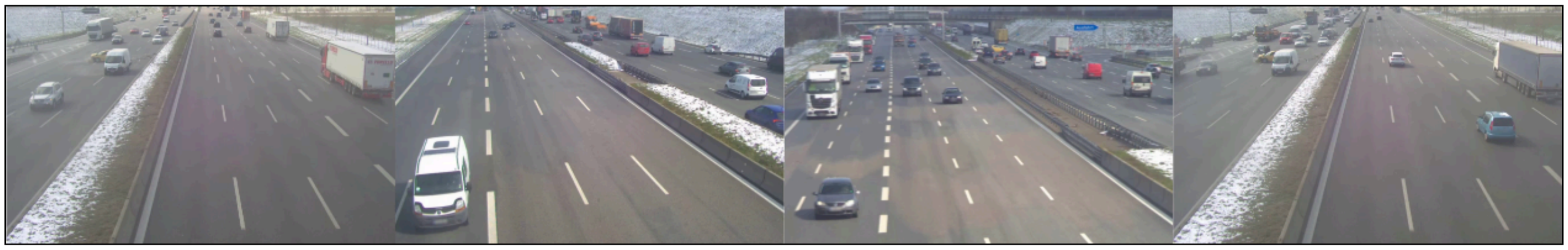

Figure 1. **VLM analysis results.** TUMTraf-Accid3nD scene images flagged as novel by the method of [4]. We note that the vehicle accident was successfully identified as novel by multiple cameras.

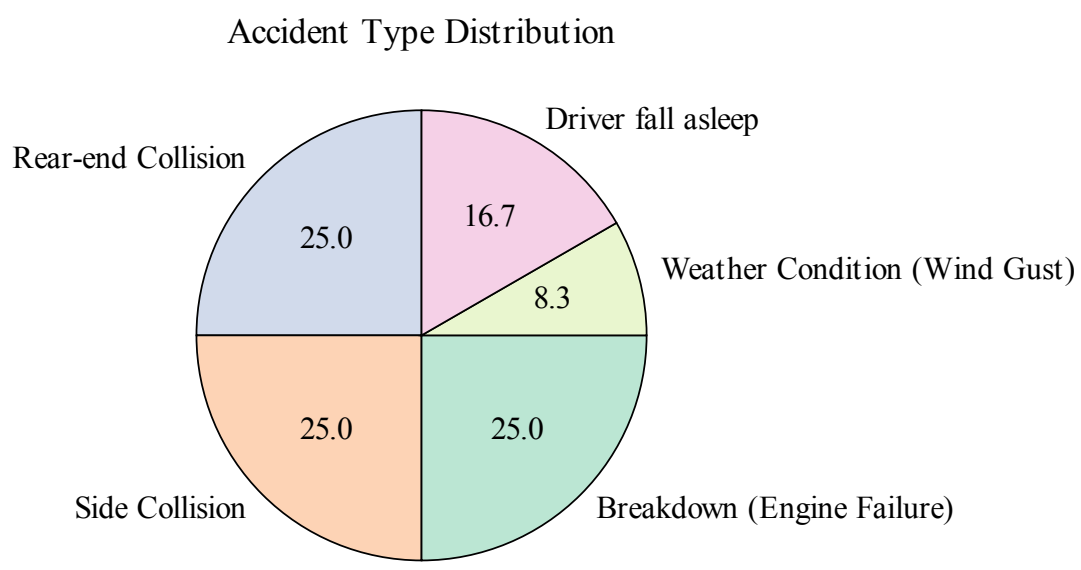

Figure 18. **Distribution of accident types.**

## D. Benchmarking Experiments

### D.1. Experimental Setup

Benchmarking experiments utilized high-performance NVIDIA GPUs (RTX 3090) and frameworks such as PyTorch for object detection and tracking. Key evaluation metrics include the mean Average Precision (mAP) for detection, F1-scores for classification tasks, and Multi-Object Tracking Accuracy (MOTA) for tracking performance. All benchmarks followed a standardized protocol to ensure reproducibility, including fixed dataset splits and uniform preprocessing pipelines. The experimental setup was validated to provide reliable baselines for future research.

### D.2. Supported Tasks

The dataset supports a range of tasks critical to autonomous driving, including 2D object detection, 2D instance segmentation, 3D object detection, object tracking, sensor data fusion, trajectory prediction, and accident detection. Additionally, its multi-modal nature enables advanced use cases such as risk assessment, scene reconstruction, and cooperative perception. This allows researchers to address fundamental challenges in safety-critical scenarios.

### D.3. Experimental Analysis

Performance varied significantly across object classes and accident scenarios. Cars, the most frequent object type, achieved the highest detection precision for accident detection, benefiting from abundant training samples. Trucks and motorcycles, while less frequent, showed lower precision due to challenges like size variability and partial occlusions. Scenarios involving sudden stops or lane changes exhibited higher accuracy compared to subtle incidents like slow drifts. This per-class breakdown informs researchers about class-specific challenges, guiding improvements in model architectures and dataset-balancing strategies.

### D.4. Vision-Language Model Efficacy

As a demonstration of VLM analysis on the TUMTraf-Accid3nD benchmark, we apply the novelty detection method of [4] to the dataset. Of the 11,924 images in the test split, 12 are flagged as most novel, with 6 of these images depicting accidents and others depicting novel events such as stopped traffic or unusually patterned vans. This method prioritizes finding events unlike others in the dataset, so the relatively high frame rate challenges the method. This can be overcome by simple downsampling of the dataset. Interestingly, the method is able to pick up on novel accident scenes from each camera view. Example figures from this novel set are displayed in Fig. 1. We note that for future research, binary annotation of accident vs. non-accident scenes can allow for training and quantitative evaluation of this novelty detection method over the dataset.

## E. Accident Detection Methodology

### E.1. Rule-Based Detection

Our rule-based detection module uses a set of pre-defined thresholds to identify accidents. For example, a vehicle decelerating at a rate exceeding 5 $m/s^2$ or experiencing a sudden trajectory deviation of over 30° triggers an accident flag. These rules are grounded in real-world traffic data and validated through extensive simulations. Pseudocode and mathematical formulas outline the detection logic, providing transparency and replicability for researchers.

### E.2. Learning-Based Detection

The YOLOv8 model [12], fine-tuned on our custom dataset, achieved a precision of 80% in detecting accident events. We used a balanced dataset split to train the model to mitigate the effects of class imbalance and to improve the detection sensitivity for less frequent accident types like side collisions. Hyperparameter tuning, such as adjusting learning rates and batch sizes, further optimized model performance. A detailed analysis of the model's sensitivity to dataset size and class distribution reveals its robustness in real-world scenarios, offering a reliable benchmark for future development.

## F. Applications and Broader Implications

The dataset supports autonomous driving applications, particularly in safety-critical scenarios.

### F.1. Autonomous Driving Use Cases

The dataset empowers autonomous driving systems by enhancing their ability to handle safety-critical situations. It supports tasks like real-time accident detection and trajectory prediction, enabling proactive measures like emergency braking or lane adjustments. For instance, recognizing abrupt braking patterns allows autonomous systems to react swiftly, minimizing collision risks.

### F.2. Infrastructure-Based Applications

Beyond vehicle systems, the dataset has significant implications for Intelligent Transportation Systems (ITS). V2X communication applications use this data to send timely alerts to nearby vehicles and emergency services, reducing response times during accidents. These interventions use sensor data fusion to ensure precise and actionable insights for infrastructure-based safety measures.

### F.3. Research Opportunities

The dataset unlocks avenues for advancing multi-modal learning, cooperative perception, and predictive modeling. Research can explore scene graph generation to map intricate relationships between objects or delve into accident anticipation models for preemptive interventions. These areas represent cutting-edge challenges that can reshape safety mechanisms in autonomous systems and transportation.

## G. Limitations and Future Directions

### G.1. Dataset Limitations

The dataset primarily focuses on highways, with limited representation of urban settings and adverse weather conditions like snow. Additionally, the scarcity of nighttime accident samples introduces biases that may impact model generalizability to diverse environments.

### G.2. Future Expansion

Plans for expansion include capturing data in urban environments, incorporating additional sensor modalities such as 4D radar and thermal imaging for low-light scenarios, and increasing the diversity of accident types. These improvements aim to bridge current gaps, making the dataset more versatile for varied applications and conditions. Our goal is also to deploy the accident detection pipeline in real-world settings to notify traffic participants in real-time. In future work, we will also explore accident video diffusion approaches for generating realistic accident scenarios, enabling better training and evaluation of autonomous driving models in rare and safety-critical situations.

## H. Development Kit Overview

The dataset is licensed under the Creative Commons License (CC BY-NC-SA 4.0) and is accompanied by a development kit comprising tools for annotation correction, model evaluation, and visualization. Pre-trained models and modular scripts are provided to simplify integration, enabling researchers to focus on experimentation and innovation.